\pdfoutput=1
\documentclass[11pt]{article}

\usepackage[final]{acl}

\usepackage{times}
\usepackage{latexsym}

\usepackage[T1]{fontenc}
\usepackage[utf8]{inputenc}

\usepackage{microtype}

\usepackage{inconsolata}

\usepackage{graphicx}

\usepackage{booktabs}
\usepackage{caption}
\usepackage{adjustbox}
\usepackage{multirow}
\usepackage{amssymb}
\usepackage{enumitem}
\usepackage{tabularx}
\usepackage{algorithm}
\usepackage{algpseudocode}
\usepackage{amsmath}
\usepackage{rotating}
\usepackage{hyperref}

\title{Detecting Latin in Historical Books with Large Language Models: \\ A Multimodal Benchmark}

 \author{
Yu Wu\thanks{These authors contributed equally.} \and
Ke Shu\footnotemark[1] \and
Jonas Fischer\footnotemark[1] \and \\
\textbf{Lidia Pivovarova} \and \textbf{David Rosson} \and \textbf{Eetu Mäkelä} \and \textbf{Mikko Tolonen} \\
University of Helsinki \\
\texttt{\{firstname.lastname\}@helsinki.fi}
}

\begin{document}
\maketitle
\begin{abstract}

%The paper proposes a novel task of extracting low-resourced language (Latin) fragments from a complex context (historical books predominantly in another language). For this task, we annotate a dataset of 724 pages and develop an evaluation method that caters to the complexities of the task and data. We run systematic benchmarking experiments using large textual, visual, and multimodal foundational models. Finally, we conduct both quantitative and qualitative evaluations and error analysis. 
This paper presents a novel task of extracting low-resourced and noisy Latin fragments from mixed-language historical documents with varied layouts. We benchmark and evaluate the performance of large foundation models against a multimodal dataset of 724 annotated pages. The results demonstrate that reliable Latin detection with contemporary zero-shot models is achievable, yet these models lack a functional comprehension of Latin. This study establishes a comprehensive baseline for processing Latin within mixed-language corpora, supporting quantitative analysis in intellectual history and historical linguistics. Both the dataset and code are available at \url{https://github.com/COMHIS/EACL26-detect-latin}.

\end{abstract}

\section{Introduction}

Accurate language identification at a granular level within historical documents is a key component to the study of the early modern period at scale. Latin, as the primary written language of Western Europe for more than a millennium, has a unique position, gradually ceding to vernaculars at varying paces across regions and genres~\citep{marjanen2025book}. Throughout this transition, Latin fragments frequently appeared within predominantly vernacular texts, in quotations, specialist terminology, and instances of code-switching. Automated extraction of diverse Latin uses in context from historical corpora is crucial to studying language evolution, the interplay between classical and modern thought, and the dissemination of ideas~\cite{sprugnoli-etal-2024-overview,gorovaia-etal-2024-sui,perrone2021lexical,burns2021profiling}. The underlying research interest behind this paper is thus to enable quantitative, fragment-level measurement of Latin’s presence in 18th-century British print, so historians and linguists can trace vernacularization across time and genres. However, this task poses challenges due to wide variations in Latin usage, scripts, complex page layouts, and inconsistent print and scan quality in historical book databases.

This study focuses on detecting instances of Latin within the \textit{Eighteenth Century Collection Online} (ECCO)~\cite{tolonen2022anatomy} corpus using both book page images and the corresponding text extracted by Optical Character Recognition (OCR). The lack of an existing dataset specifically designed for multimodal and code-mixed Latin detection motivated us to create an annotated dataset for this purpose. Our dataset contains 724 pages sampled from historical documents, validated by specialists in 18th-century publishing culture to represent diverse use cases, with our novel, fine-grained typology of 12 categories. In future work at ECCO scale, our focus on reliable detection and page-level quantification will support temporal trend analyses (e.g., decade-by-decade decline, genre-specific persistence, bilingual presentation with English glossing) and reconstructing the footprint of learned discourse in print culture. While we focus on Latin due to its aforementioned importance for historical study, our fully benchmarked, manually annotated scenario provides a solid template for extending the method to other languages as well.

% and has been validated by specialists in 18th-century publishing culture.

Given all the complex nature of the task, %and factors ranging from Latin being a relatively low resource language and OCR noise to varied print layouts and the widespread adaption of Latin vocabulary into other languages, including English, 
we explore the capabilities of modern Large Language Models (LLMs), including multimodal models (MLLMs) for this task. The way these models handle contextual information, recognize patterns in noisy data, and integrate textual cues with visual layout information has been found to help disambiguate text and languages in historical documents~\cite{luo2024layoutllm, boros-etal-2024-post, kanerva2025ocr, xie2025multimodal} compared to traditional Natural Language Processing (NLP) methods. %Their capacity for nuanced understanding makes them promising candidates for tackling the specific challenges of identifying embedded Latin, which often requires sensitivity to both local textual features and broader document context.
Our investigation exploits the new dataset to test a number of state-of-the-art models and finds that reliable Latin detection in such challenging historical material is achievable. The benchmarking of different model architectures provides insight into their strengths and weaknesses when faced with the complexities inherent in the data. This work establishes a strong baseline for a novel NLP task and highlights the need for more modality and semantic-aware approaches, as well as robust evaluation frameworks in historical text analysis.

The main contributions of this article are:

\begin{itemize}[topsep=2pt, partopsep=2pt]
\itemsep -0.4em
\item Defining the novel multimodal task of Latin detection in historical documents.
\item Creating a new benchmark dataset of diverse Latin usage in 18th-century books.
\item Developing a robust evaluation framework tailored to the multimodal challenges of the task.
\item Providing a practical Latin detection pipeline that enables large-scale downstream applications in historical research.
\item Systematically benchmarking contemporary LLMs for this task.

% \item Defining the task of Latin detection in historical documents, an understudied multimodal case of language detection.
% \item Creating an expert-annotated dataset from 18th-century books, capturing diverse and challenging examples of Latin usage.
% \item Developing an evaluation framework for this task, considering challenges from both textual and visual modalities.
% \item Delivering a practical Latin detection pipeline, demonstrating its readiness for downstream applications in historical research.
\end{itemize}

\section{Problem Definition}
We define the task of \textit{Latin language detection in historical documents} as a two-stage classification and extraction problem, where the input consists of a scanned page image and/or its OCR transcription. The task is to automatically detect whether any segments in the text are written in Latin, and if so, to extract text of those specific segments.

Formally, given a document page \( D \), let \( I_D \) denote its image and \( T_D \) denote its OCR-processed text. A system must perform the following two subtasks:

\begin{itemize}[topsep=2pt, partopsep=2pt]
  \itemsep -0.2em
  \item \textbf{Task 1 (Page-level Latin Detection):} Predict a binary label \( y_D \in \{0,1\} \), where \( y_D = 1 \) indicates that the page contains at least one segment in Latin, and \( y_D = 0 \) otherwise.
  \item \textbf{Task 2 (Latin Segment Extraction):} If \( y_D = 1 \), extract a list of text spans \( S_D = [s_1, s_2, \ldots, s_n] \), where each \( s_i \in T_D \) is a contiguous Latin segment string.
\end{itemize}

% The reason for specifying the task in two stages is mainly due to evaluation. As our final aim is to evaluate how well Latin is identified at a high level of granularity across different layouts, we measure task 2 in terms of per-page precision and recall. However, this type of measurement does not adequately cater to cases where Latin is not present on a page. Thus, task 1 has been designed to give us this more general information on e.g., whether the models are excessive in detecting Latin where there is none. Requiring segment-level output as strings rather than image regions aligns better with tasks most MLLMs are trained on, and enables simpler performance comparisons between the input modalities. More details on the metrics used will be provided in Section~\ref {sec:evaluation}.

We structure our problem into two tasks for a more comprehensive evaluation. The core challenge, Task 2, is the fine-grained extraction of Latin text segments, evaluated using per-page token precision and recall. Since these metrics inadequately handle pages with no Latin, we introduce Task 1, a page-level binary classification, to assess performance on these non-Latin instances specifically. We require extracted segments as strings rather than image regions, as this output format better aligns with the capabilities of most MLLMs and enables a simpler, more direct comparison across different input modalities. Detailed definitions of our metrics are provided in Section~\ref{sec:evaluation}.

\section{Related work}

\vspace{-0.5em}
\paragraph{Latin in NLP} Given its historical importance, Latin has attracted considerable attention within the NLP community \cite[e.g.,][]{sprugnoli-etal-2024-overview,schulz-keller-2016-code,gorovaia-etal-2024-sui,perrone2021lexical,burns2021profiling}, though much of this research has centered on small, clean corpora of ancient literary texts. While some recent studies have ventured into Early Modern mixed-language documents \citep{stussi-strobel-2024-part,volk-etal-2024-llm}, these also predominantly rely on manually curated and annotated data. In contrast, our work focuses on the foundational task of Latin \textit{discovery}: detecting Latin within extensive, unedited, and noisy digitized collections like ECCO \cite{tolonen2022anatomy}. This computational approach aims to detect Latin in a vast corpora, while the identified fragments can subsequently be analyzed using a range of established NLP tools developed for classical languages \citep{johnson-etal-2021-classical,burns2023latincy,straka-strakova-2020-udpipe,kupari-etal-2024-improving}.

% Given its importance, Latin has attracted much attention from the NLP community \cite[among others]{sprugnoli-etal-2024-overview,schulz-keller-2016-code,gorovaia-etal-2024-sui,perrone2021lexical,burns2021profiling}. Most of these work deals with small, clean corpora compiled of ancient literary texts. Few works deal with Early Modern mixed-language text \citep{stussi-strobel-2024-part,volk-etal-2024-llm}, and they also mainly use clean, manually annotated data.

% Our work is different because we deal with Latin \textit{discovery} rather than analysis, and our data consists of whole books, digitized but unedited and understudied. The exact amount and nature of Latin within ECCO collection are unknown, thus our work contributes to ongoing historical research. On the other hand, once Latin fragments are detected, a number of matured tools would be applicable for further analysis \citep{johnson-etal-2021-classical,burns2023latincy,straka-strakova-2020-udpipe,kupari-etal-2024-improving}.

\vspace{-0.5em}
\paragraph{Code-mixed Language Detection}From a methodology perspective, identifying Latin segments within historical publications is a code-mixed language detection task \citep{aguilar-etal-2020-lince}. While extensive research in this area has focused on contemporary informal texts \citep{barman-etal-2014-code, zhang-etal-2018-fast-compact}, its application to historical documents, with challenges like archaic syntax, lexicon, and spelling, has been less explored \citep{schulz-keller-2016-code, volk-etal-2022-nunc}. Detecting classical languages in these complex historical contexts has traditionally involved rule-based systems and supervised machine learning approaches, notably Conditional Random Fields (CRFs) \citep{schulz-keller-2016-code, sterner-teufel-2023-tongueswitcher, volk-etal-2022-nunc}. Alongside these, robust statistical tools like Lingua \citep{pemistahl2021lingua} offer effective general language identification with support for mixed language. Given the recognized potential of modern LLMs to navigate linguistic nuances and noisy data, our work investigates their capacity to enhance detection performance.

% Code-mixed language detection, or code-switching detection, involves the identification of language boundaries within texts that contain multiple languages \citep{aguilar-etal-2020-lince}. Previous research has primarily examined informal contemporary contexts such as spoken dialogues and social media, as highlighted by \citet{barman-etal-2014-code} and \citet{zhang-etal-2018-fast-compact}. More recently, there has been a growing interest in extending these methods to historical documents, which present unique linguistic challenges such as archaic syntax, extensive spelling variations, and scarce annotated data \cite{schulz-keller-2016-code,volk-etal-2022-nunc}.

% Latin frequently appears in historical multi-lingual documents due to its historical significance \citep{schulz-keller-2016-code}. Early historical code-switching detection research relied mainly on rule-based systems using dictionaries and morphological analysers. However, these methods often struggled with accurately identifying ambiguous or previously unseen words \citep{sterner-teufel-2023-tongueswitcher}. To address this issue, more recent research has incorporated supervised machine learning approaches, notably Conditional Random Fields (CRFs) that use linguistic properties, such as character and word n-grams, context windows, and morphological features~\citep{schulz-keller-2016-code, volk-etal-2022-nunc}.

\vspace{-0.5em}
\paragraph{LLMs for Historical Documents} Recent LLMs, particularly Multimodal variants (MLLMs), have shown considerable potential in historical document analysis, demonstrating top performance in tasks like OCR, named entity recognition, and general document understanding from historical sources \citep{Qwen2.5-VL, luo2024layoutllm, boros-etal-2024-post, kanerva2025ocr, backer2025bootstrapping, xie2025multimodal}, and in assessing general historical knowledge \citep{hauser2024large}. Despite these advancements, a significant gap persists for more specialized, complex applications. Specifically, there is a notable lack of dedicated benchmarks and systematic exploration for the fine-grained, page-level multimodal detection and extraction of embedded secondary languages (e.g., Latin) \citep{aguilar-etal-2020-lince, guzman2017metrics}. This task is demanding due to noisy scans from historical archives, diachronic language context, and orthographic variation \citep{volk-etal-2022-nunc}. Our work contributes to this underexplored area by introducing a systematic evaluation methodology designed to be scalable also to other languages. 
% todo: add new historical ocr papers

% Further advancements came from neural models, especially those employing pre-trained multilingual embeddings such as multilingual BERT and GPT, which capture nuanced contextual information \citep{sterner-teufel-2023-tongueswitcher}. These neural techniques utilize context-rich embeddings, language modelling, and sequence labelling, often supplemented with classical supervised classification methods like Support Vector Machines (SVM) and decision trees \citep{sterner-teufel-2023-tongueswitcher,zhang-etal-2018-fast-compact}.

% Nevertheless, as emphasized by \citet{volk-etal-2022-nunc}, the inherent orthographic variability in historical texts necessitates specialized preprocessing and normalization, along with careful integration of linguistic resources. Additionally, \citet{aguilar-etal-2020-lince} and \citet{guzman2017metrics} have identified
% the notable absence of standardized benchmarks tailored specifically for historical multilingual contexts, particularly for Latin-English language pairs.

\section{Dataset} 

\begin{figure}[t!]
    \centering
    %  trim={<left> <lower> <right> <upper>}
    \includegraphics[width=\linewidth, clip]{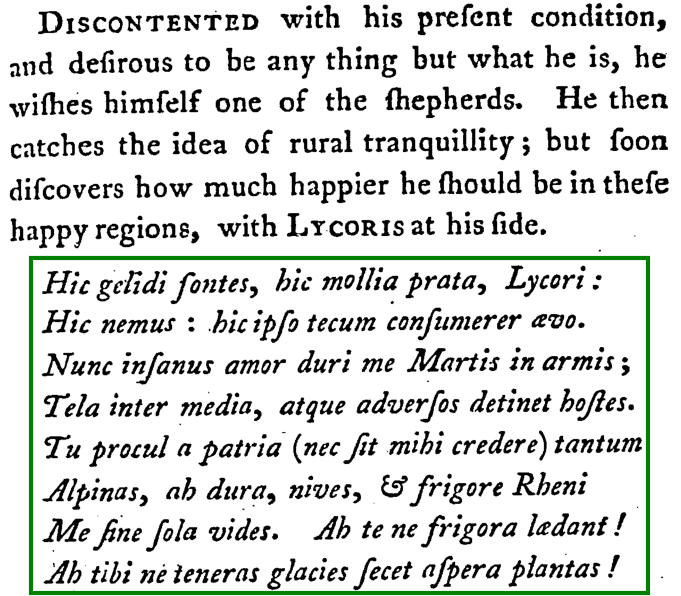}
    \caption{An example of an annotated Latin fragment and surrounding context.}
    \label{fig:example-page1}
    \vspace{-1em}
\end{figure}
\subsection{Sampling and Annotations}
Our approach to dataset construction began with a targeted sampling strategy to identify pages with a high likelihood of containing Latin text. We queried the Reception Reader database~\cite{rosson2023reception}, which indexed text reuses across the ECCO corpus using noise-resistant detection methods. From this, we randomly selected 800 reuse instances where one book was cataloged as Latin and the other as non-Latin. To ensure broad representation and reflect the diversity of the ECCO collection (approximately 200,000 books), each sampled page was drawn from a different book, covering varied publication dates and topics. However, ECCO's language metadata is book-level, meaning that ``Latin'' books often contain significant non-Latin text like English introductions. Also, the reuse offsets only mark the textual overlap without specifying the language of the text segment.

These pages were then manually annotated. Annotators were tasked with drawing bounding boxes around all Latin fragments on the page images (see example in Figure~\ref{fig:example-page1}). These visual annotations are later reliably mapped to text offsets (locations within a string) using ECCO’s OCR positional data for ground truth text extraction. Our annotation guidelines stated marking all instances of Latin text semantically used as Latin. This included single Latin words if presented with explanations in a dictionary, as well as Latin found in headlines, editorial annotations, or footnotes. %Conversely, to ensure clarity and consistency, in-line Roman/Latin names (of people, places, plants) and jargon were not specifically annotated as Latin but were treated as part of the surrounding language, as well as abbreviations, such as 'etc'.

The annotation environment was Label Studio~\cite{LabelStudio}. The primary annotation was performed by three scholars familiar with Latin. Following this, an expert in historical texts meticulously reviewed and validated all annotations to ensure accuracy and consistency. During the process, the annotators veto the pages with incompletely OCR-transcribed regions.

\subsection{Dataset Characteristics}
\label{sec:categories}

In total, 724 pages were annotated, with 594 identified as containing Latin. An expected finding during annotation was the frequent presence of other languages, such as French, German, and Greek, highlighting the dataset's challenging multilingual nature beyond simple Latin-English code-mixing.

To contextualize model performance and to better understand the dataset's composition, we divided the annotated Latin segments into 12 language integration categories. Each category represents a specific way in which Latin is used in 18th-century British books and how it relates to English-language text. Depending on their content, all Latin text segments were assigned to one or multiple categories, with some frequently appearing together (e.g., footnotes and code-switching), while others are mostly exclusive (e.g., bilingual). This novel categorization enables a fine-grained analysis of both historical linguistic practices and the performance of our Latin detection approach within different contexts of language integration. %The following list defines the categories used:
\begin{table}[ht!]
\centering
\begin{tabular}{r|l|r}
& \textbf{Category} & \textbf{Count} \\
\hline
1. & Direct Quote & 258 \\
2. & Independent & 196 \\
3. & Footnote & 191 \\
4. & Code-switching & 100 \\
5. & Bilingual & 55 \\
6. & Emblematic & 34 \\
7. & Indices and Catalogs & 30 \\
8. & Legal & 30 \\
9. & Ecclesiastical & 23 \\
10. & Tables and Charts & 11 \\
11. & Dictionary & 9 \\
12. & Side-note & 8 \\
\hline
\end{tabular}
\caption{Page counts by Latin segment category.}
\label{tab:annotation-categories}
\vspace{-1em}
\end{table}

Table~\ref{tab:annotation-categories} shows all the categories and the frequency of each annotation category within our dataset. The full definitions of 12 categories are listed in Appendix~\ref{appendix:latin_categories}, and categorized segment examples are shown in Appendix~\ref{appendix:append_data}. 
%Figures~\ref{fig:example-page2} and~\ref{fig:example-page3} in the 

%This form of page-level annotation reduces the need for costly instance-level labeling, which is particularly challenging for Latin due to expertise requirements and high annotation volume. Moreover, it supports context-rich evaluation aligned with the page-level structural nature of our detection task inputs.

\section{Evaluation Setting}
\label{sec:evaluation}

%To systematically evaluate model performance on Latin segment detection, a two-part evaluation strategy corresponding to the two sub-tasks defined earlier is adopted. 

\subsection{OCR Post-Correction and Normalization}
\label{sec:ocr_setting}

% We noticed that the OCR performance of modern LLMs is much higher than that of the models used to OCR the ECCO collection. Thus, in our result models that take an image as input output text is much cleaner than the noisy text from the ground truth. For that reason, evaluating using strict string similarity would punish these models for outputting cleaner text. Therefore, it makes sense to correct OCR errors in the ground truth. However, this makes it difficult to evaluate text-based models that reproduce OCR noise for the input. For this reason, we post-corrected both ground truth and input pages, that was the only way to ensure unified evaluation for both text-based and image-based model. For that end we used GPT4-o1~\cite{achiam2023gpt} model and a prompt from~\citet{kanerva2025ocr} to get the clean strings of Latin text and page text. % todo: more explanations on the correction process?
% Note that despite postcorrection, the significant amount of OCR noise still remains in the data, matching model output and ground truth remains a non-trivial task (see Section~\ref{sec:evaluation}).

Evaluating models on the ECCO corpus is complicated by significant OCR quality discrepancies: modern models with vision capabilities often produce cleaner text than ECCO's original OCR, while text-based models may or may not replicate the noise in their input. Such differences make direct string-based comparisons problematic and distort evaluation. To ensure meaningful assessment across all model types, we post-correct both the ground-truth Latin segments and the full input page texts. This OCR post-correction is performed using the OpenAI o1 model~\cite{jaech2024openai} with a specialized prompt from~\cite{kanerva2025ocr}. 

Even after the post-correction, residual noise and other variation still remain in the data. Thus, for token-based evaluation, we apply a more traditional rule-based preprocessing pipeline to both predicted and reference strings. This deterministic pipeline, informed by our extensive experience with OCR data and domain-specific knowledge, targets common superficial textual variations and ensures a fair alignment. The pipeline includes Unicode normalization, ligature replacement, lowercasing, digit removal, de-hyphenation, and punctuation stripping. Subsequent to these cleaning operations, the strings are tokenized into word-level units. More details on the processing steps are presented in Appendix~\ref{appendix:preprocess}.%This thorough standardization is crucial for eliminating noisy tokens to ensure a fair alignment in the token-level language comparison. More details of the processing steps could be found in the Appendix. % details should be in supply.
% todo: add more justifications and details on the processing

\subsection{Metrics}
\label{sec:metrics}
The goal of Task 1 is to detect whether a page has Latin on it. We measure this by reporting precision, recall, and F1 score in percentage, along with the F1 score for non-Latin pages to ensure balanced evaluation.  % This reflects the detection-oriented nature of the task. %To better quantify the impact of false positives—pages incorrectly classified as containing Latin—we additionally compute the proportion of falsely predicted Latin tokens relative to the total number of tokens on the page. This provides a fine-grained measure of spurious content detection. 
To evaluate the Latin segment extraction performance in Task 2, we calculate precision, recall, and F1 score in percentage based on token-level matches between model predictions and the ground truth of the page. 
A fuzzy matching mechanism is applied to pair predicted and reference tokens one by one. A match is considered valid if the token-level edit distance is not larger than a tunable threshold proportion $\theta$ compared to the ground-truth token length. This approach provides a more flexible and robust evaluation than exact token matching by tolerating minor textual differences at the token level, such as lexical variations and OCR-induced distortions. The pseudocode of the fuzzy matching algorithm is shown in the Appendix~\ref{appendix:fuzzy_alg}. Overall metrics are averaged across the full evaluation set.

% This paper explores both visual and textual approaches to Latin detection. Correspondently, the evaluation can be done using either image or text. Image-based evaluation would measure whether automatically found Latin areas correspond to manually annotated rectangles. However, in those rectangles much of the area is taken by blank space, which may distort evaluation. Thus, we decide to use text-based evaluation. However, we noticed that the OCR performance of modern LLMs is much higher that those of the models used to OCR the ECCO collection. In a multimodal setting, where a model that takes image as an input and outputs text, this output text is much cleaner than the noisy text from the ground truth. For that reason, evaluating using strict string similarity would punish these models for outputting cleaner text. Thus, to obtain most informative evaluation scores we performed several clean-up and alignment steps.

% Yu:

%OCR-noise and postcorrection effect

% All punctuation marks and numbers are excluded from evaluation. In our annotation process, larger text fragments annotated as Latin can include some extra numbers, such as line numbers. In other cases numbers are parts of the annotated text, and sometimes it is difficult to distinguish between these two cases. 

% punctuation and numbers are excluded from evaluation (they are sometimes included into Latin annotation sometimes not 

% historical symbols, ligatures  ?

% Shortness in length means very few OCR errors can significantly impact our scores 

% string alignment, 1-gram, 4-gram

\section{Latin Extraction Pipeline with LLMs}
% multimodal approaches considered, with unified text segments output

%We design a unified, prompt-based pipeline to extract Latin text segments from historical document pages using large foundation models. The core idea is to frame both subtasks as a single instruction-following task, enabling consistent usage off-the-shelf across diverse input modalities. %, including text-only, vision-only, and combined ones. 

%\paragraph{Large Language Models with Multimodal Capabilities}
%LLMs have demonstrated strong generalization ability across a wide range of natural language understanding and generation tasks. Recent advances have extended this paradigm to support various input modalities beyond plain text, leading to models capable of processing not only textual but also visual content. 
%In our pipeline setting,
%We make use of such instruction-following LLMs that accept either text-only, image-only, or combined inputs, and return structured text outputs in response to a unified prompt. This allows us to reuse the same extraction pipeline across models with different input modalities.

Our evaluation investigates the application of general instruction-following LLMs, particularly multimodal variants, for Latin segment extraction from historical documents. We propose a unified, prompt-based pipeline designed to be both practical for real-world deployment and robust for systematically and fairly evaluating the capabilities of diverse LLMs on this task. 

\vspace{-0.5em}
\paragraph{Unified Prompting Strategy}
We employ a minimal, high-level instructional prompt designed to elicit responses that inherently address both sub-tasks within one simply formatted output. This approach simplifies interaction with the models and the subsequent processing of their outputs, thereby contributing to the overall ease of application.

This unified prompt asks the LLM to extract all Latin segments to a list, without further instructions. The distinction in our experiments lies solely in the input provided to this consistent prompt, where the specific prompts are shown in the Appendix~\ref{appendix:prompts}:

%Specifically, the prompt instructs the LLM to return its findings as a JSON list. If no Latin script is detected on the page, the LLM is expected to return an empty list. The presence of a non-empty list directly indicates that Latin script has been identified, and the elements within the list constitute the extracted Latin segments. This encoding of both tasks within the requested output structure streamlines the subsequent processing across different input modalities, LLM architectures and various layouts.

\begin{itemize}[topsep=1pt, partopsep=1pt]
    \itemsep-0.4em
    \item \textbf{Text-only:} The OCR-extracted and post-corrected text, appended to the prompt.
    \item \textbf{Image-only:} The page image, with the prompt guiding it to the visual signal.
    \item \textbf{Multimodal:} Both the scanned page image and the corrected OCR text are included.
\end{itemize}
%This unified prompting strategy, coupled with the varying input modalities, allows for a direct and controlled comparison of the impact of different information sources on the LLM's ability to perform Latin script detection and extraction.

\vspace{-0.5em}
\paragraph{Structured Output and Postprocessing}
The LLMs are instructed to output their predictions as a list of Latin segments, which directly corresponds to the output requirement for Task 2. 
The presence of a non-empty list implicitly indicates the presence of Latin script on the page (Task 1, \(y_D = 1\)), while an empty list indicates its absence (\(y_D = 0\)). 

\vspace{-0.5em}
\paragraph{Model-Agnostic Compatibility}
Because the method does not rely on any model-specific architecture or training, it can be directly applied to a wide range of general-purpose foundation language models. This makes the approach particularly suitable for scalable deployment across large historical corpora with variable OCR quality and image-text alignment conditions.

\section{Experiments}
%Ke+Yu
% We conduct a series of experiments to evaluate the effectiveness of our instruction-based extraction pipeline on historical Latin document data. This section is structured as follows: we first introduce the experiment setup, %which includes the large language models applied in our experiments and the implementation details such as the inference platform and prompts. 
% and then present quantitative and qualitative results. We assess both the overall utility of general-purpose language models for this task and the impact of different model designs on extraction accuracy.

\subsection{Experiment Setup}
\paragraph{Model Selection}
To explore how modern instruction-tuned language models handle the new task of Latin segment detection and extraction in noisy multimodal historical documents, we benchmark a representative suite of LLMs across modalities, scales, and architectures.
The model selection follows three guiding principles:
(i) in the absence of dedicated multimodal benchmarks for historical language understanding in documents, we refer to leaderboard performance on \textbf{DocVQA} \citep{tito2021document} and comprehensive open evaluations such as \textbf{OpenCompass} \citep{2023opencompass} and \textbf{MMMU} \citep{yue2024mmmu}; %MMLU also, but it is in text QA
(ii) we prioritize lightweight to medium-scale models (7B-72B) to better reflect realistic research use cases in historical academic and limited-resource scenarios. Specifically, the selected models include: % (additionally, we evaluate further models detailed in the Appendix~\ref{appendix:other_llms}):
\begin{itemize}[topsep=2pt, partopsep=2pt]
  \itemsep-0.4em 
  \item \textbf{GPT-4.1} \citep{achiam2023gpt}: proprietary frontier MLLM accessed via the OpenAI API, included as a strong reference point for the performance without explicit thinking mode. 
  \item \textbf{Qwen2.5-VL series} (72B, 32B, 7B) \citep{Qwen2.5-VL}: open-source flagship MLLMs with strong document understanding and visual grounding. We include both Vision-Language and text-only instruction-tuned variants to disentangle the multimodal inputs. 
  \item \textbf{Qwen3 series} (235B-A22B, 30B-A3B, 32B, 14B, 4B) \citep{qwen3technicalreport}: latest generation of Qwen text models with redesigned architecture and built-in \emph{thinking} mode. We evaluate them to probe how reasoning-augmented LLMs transfer to the historical language task.
  \item \textbf{DeepSeek-R1 variants} (original and distilled) \citep{guo2025deepseek}: pioneering reasoning-centric LLMs. We use the original large R1 to gauge frontier open-source LLMs, and distilled versions (based on Llama-3.3-70B and Qwen2.5-32B) to examine transfer of reasoning signals into smaller models.
  \item \textbf{InternVL3 series} (38B, 14B, 8B) \citep{zhu2025internvl3}: top academic MLLMs with a two-stage visual encoder integrated into a transformer backbone. 
  \item \textbf{Gemma3} (27B) \citep{team2025gemma}: new open-source MLLM from Google, optimized for efficiency and multilingual capability. 
\end{itemize}

%Each model, if supported, is evaluated under three input configurations:\emph{text‑only}, \emph{image‑only} and \emph{multimodal} to assess how different modalities contribute to detection and extraction performance.

\vspace{-0.5em}
\paragraph{Baseline} We employ \textbf{Lingua} \citep{pemistahl2021lingua}, a statistical language identifier based on character n‑gram modelling, and the only off-the-shelf tool we found that supports token-level Latin detection in mixed-language text. While not designed for noisy OCR, it offers a practical baseline to contextualize the difficulty of our task and the potential advantages and drawbacks of LLM-based approaches. The configuration details are in Appendix~\ref{appendix:baseline}.

\vspace{-0.5em}
\paragraph{Implementation Details}
All models except GPT-4.1 are run on the supercomputer’s AMD MI250X GPU nodes via the local \texttt{vLLM} \citep{kwon2023efficient} server, using a maximum batch size of 16 with asynchronous requests and a Ray backend. Depending on model size, we allocate 1–6 nodes, totalling about 8 GPU hours per model on average. 
We use deterministic generation with a temperature set to 0 and a fixed seed to ensure reproducibility. We limit the output token length to 20k tokens. For models supporting thinking mode, we additionally enforce a thinking budget of 15k tokens to prevent looped or unbounded reasoning traces. 
The edit-distance threshold $\theta$ is set to 0.2 based on empirical evidence, discussed further in the Appendix~\ref{appendix:theta}.

\begin{table*}[ht!]
\centering
\begin{adjustbox}{max width=\textwidth}
\setlength{\tabcolsep}{3pt}
\begin{tabular}{l|l|l|c|ccc|c|ccc}
\toprule
\multicolumn{4}{c|}{\textsc{Model Details}} & \multicolumn{4}{c|}{\textsc{Page-level (Task 1)}} & \multicolumn{3}{c}{\textsc{Token-level (Task 2)}} \\
\hline
\textbf{Model} & \textbf{Variant} & \textbf{Size} & \textbf{Mode} & \textbf{F1} & \textbf{Precision} & \textbf{Recall} & \textbf{NL. F1} & \textbf{F1} & \textbf{Precision} & \textbf{Recall} \\
\midrule
Lingua & - & - & T & 92.58 & 86.32 & \textbf{99.83} & 43.11 & 77.14 & 77.31 & 80.07 \\
\midrule
GPT-4.1 & - & - & I+T & 96.03 & 92.51 & \textbf{99.83} & 77.00 & 84.59 & 88.49 & 83.54 \\
\midrule
\multirow{3}{*}{DeepSeek-R1} & - & 671B & T & 97.53 & 95.34 & \textbf{99.83} & 87.07 & \textbf{87.00} & \textbf{89.69} & 86.17 \\
\cmidrule(lr){2-11}
 & Distill-Llama & 70B & T & 96.18 & 92.94 & 99.66 & 78.34 & 84.26 & 87.81 & 83.62 \\
 \cmidrule(lr){2-11}
& Distill-Qwen & 32B & T & 95.35 & 92.55 & 98.32 & 74.44 & 82.84 & 86.14 & 82.66 \\
\midrule
\multirow{5}{*}{Qwen3} & MoE-A22B & 235B & T & 96.90 & 94.13 & \textbf{99.83} & 80.52 & 86.54 & 89.16 & 86.00 \\
\cmidrule(lr){2-11} 
& MoE-A3B & 30B & T & 95.90 & 93.45 & 98.48 & 78.07 & 84.29 & 87.29 & 83.73 \\
\cmidrule(lr){2-11}
& - & 32B & T & 95.39 & 93.25 & 97.64 & 75.86 & 84.07 & 86.46 & 83.98 \\
\cmidrule(lr){2-11}
& - & 14B & T & 96.61 & 94.96 & 98.32 & 82.85 & 84.78 & 88.10 & 83.71 \\
\cmidrule(lr){2-11}
& - & 4B & T & 88.45 & 87.36 & 89.56 & 43.27 & 75.38 & 76.30 & 77.20 \\
\midrule
\multirow{9}{*}{Qwen2.5} & VL & 72B & I+T & 98.58 & 98.16 & 98.99 & 93.33 & 83.87 & 86.98 & 82.74 \\
& VL & 72B & I & \textbf{98.82} & 98.66 & 98.99 & \textbf{94.57} & 80.00 & 82.36 & 79.72 \\
& - & 72B & T & 97.85 & \textbf{100.00} & 95.79 & 91.23 & 80.95 & 85.31 & 79.33 \\
\cmidrule(lr){2-11}
& VL & 32B & I+T & 96.18 & 92.94 & 99.66 & 78.34 & 84.32 & 86.90 & 83.99 \\
& VL & 32B & I & 96.80 & 94.40 & 99.33 & 82.97 & 79.82 & 82.64 & 79.18 \\
& - & 32B & T & 97.26 & 99.13 & 95.45 & 88.65 & 81.57 & 84.46 & 80.99 \\
\cmidrule(lr){2-11}
& VL & 7B & I+T & 88.36 & 82.89 & 94.61 & 15.91 & 72.41 & 78.62 & 72.23 \\
& VL & 7B & I & 95.75 & 96.74 & 94.78 & 81.62 & 71.42 & 76.99 & 70.39	\\
& - & 7B & T & 91.86 & 92.49 & 91.25 & 64.18 & 60.52 & 65.78 & 64.57 \\
\midrule
\multirow{6}{*}{InternVL3} & - & 38B & I+T & 95.62 & 92.32 & 99.16 & 75.00 & 84.24 & 84.43 & \textbf{86.67} \\
& - & 38B & I & 89.51 & 87.22 & 91.92 & 43.86 & 56.53 & 61.11 & 55.61 \\
\cmidrule(lr){2-11}
& - & 14B & I+T & 94.28 & 90.42 & 98.48 & 65.70 & 81.36 & 82.63 & 83.87 \\
& - & 14B & I & 90.16 & 87.86 & 92.59 & 47.37 & 53.51 & 56.46 & 55.08 \\
\cmidrule(lr){2-11}
& - & 8B & I+T & 90.54 & 86.39 & 95.12 & 41.00 & 69.04 & 65.87 & 82.10 \\
& - & 8B & I & 90.82 & 88.50 & 93.27 & 50.88 & 60.22 & 60.30 & 65.63 \\
\midrule
\multirow{3}{*}{Gemma3} & - & 27B & I+T & 90.77 & 83.57 & 99.33 & 18.92 & 82.50 & 84.00 & 84.79 \\
& - & 27B & I & 88.42 & 82.60 & 95.12 & 12.94 & 60.03 & 62.78 & 61.68 \\
& - & 27B & T & 94.22 & 90.03 & 98.82 & 64.36 & 83.79 & 86.09 & 84.20 \\
\bottomrule
\end{tabular}
\end{adjustbox}
\caption{Experimental results on selected LLMs, compared with Lingua baseline. ``\textbf{NL. F1}'' denotes the F1 score for identifying Non-Latin pages. ``I'' and ``T'' indicate image and text input separately.}
\label{tab:results}
\vspace{-1em}
\end{table*}

\begin{figure*}[t!]
    \centering
    \includegraphics[width=\textwidth]{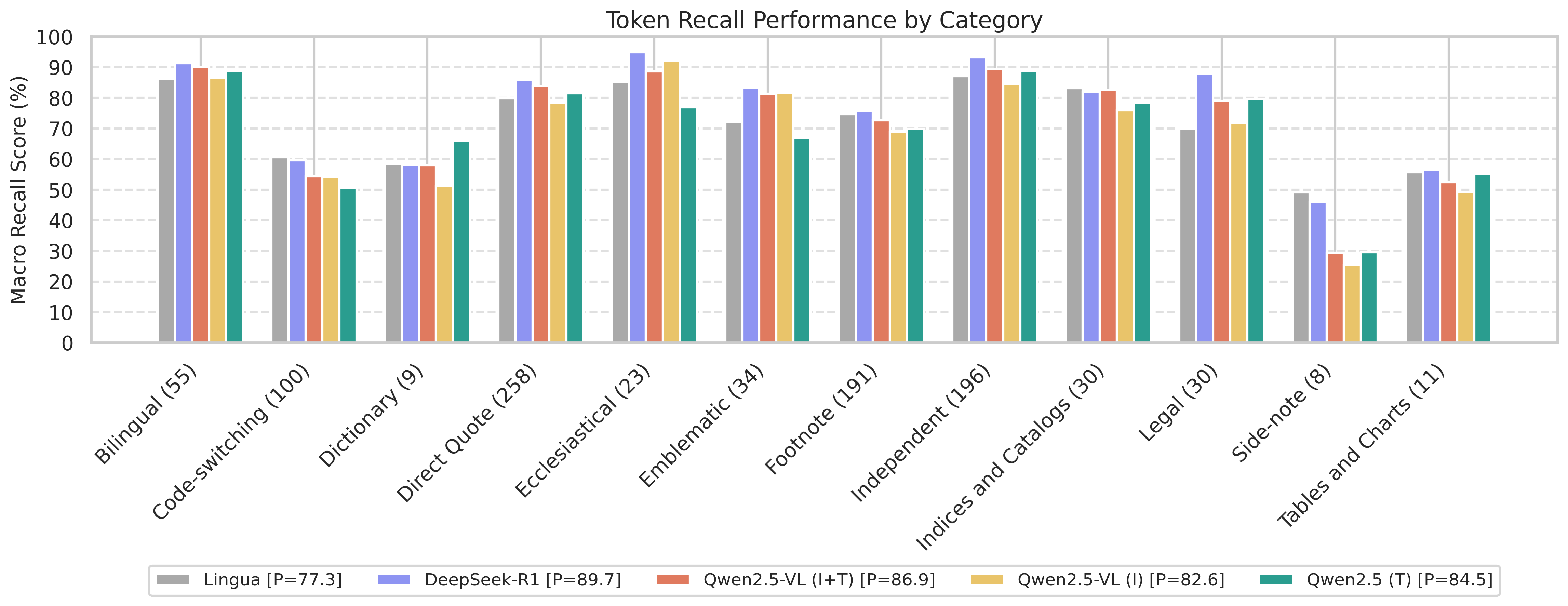}
    \caption{Macro token recall statistics on different category labels for 5 top-performing models. Qwen2.5 models are all with 32B parameters. The number of page instances with each label tagged is shown in parentheses. Values in the legend indicate each model’s token precision from Table~\ref{tab:results} to provide complementary performance context.}
    \label{fig:cat_res}
\end{figure*}

\subsection{Model Results and Analysis}
Table~\ref{tab:results} presents the overall results on the two tasks. The traditional Lingua baseline remains competitive but is consistently surpassed by recent foundation models. Notably, several large open-source LLMs such as DeepSeek-R1 and Qwen3 not only outperform Lingua but also exceed multimodal GPT-4.1 across both tasks, highlighting the rapid progress of open-source development. The different strengths of DeepSeek and Qwen2.5-VL also reveal trade-offs between tasks and metrics, suggesting that thinking trace may shift error profiles. Vision-language models further demonstrate the potential benefit of multimodal input for OCR-heavy tasks, though vision-only settings remain challenging. Overall, the open-source LLMs with thinking mode now define the frontier for this task. Future MLLMs are expected to feature more deeply integrated reasoning capabilities.

% In the page-level detection task (Task~1), Lingua achieves a strong F1 score of 92.58. In the more challenging segment extraction task (Task~2), it still attains 77.14 F1 using simple token-level identification. %This suggests that ngram-level statistical features remain robust against much OCR noise.
% Qwen2.5-VL in particular demonstrates strong document-oriented OCR capabilities, being the only model to excel even in vision-only settings. %Smaller VL models, like InternVL3-14B, can still improve over Lingua when vision is used to guide text modeling, but fail in vision-only configurations, illustrating that OCR robustness is not easy to achieve. 
% Interestingly, the DeepSeek-R1-Distilled version of Qwen does worse than plain Qwen on Task 1, but better on Task 2. Looking in more details at the numbers, the distilled version seems to be simply optimising for recall at the cost of precision on the page-level metric, while then being more accurate in the actual segment extraction.

%We state several specific findings as following:

%\paragraph{Task Comparison}
%As expected, all models perform substantially better on Task~1 than on Task~2, though model rankings remain largely consistent across tasks. This suggests that although Task~1 is structurally simpler, it is still reflective of deeper capabilities such as multilingual grounding and robustness to OCR noise. Hence, Task~1 can serve as a reliable proxy for identifying strong Latin-handling models, particularly in settings where full extraction may be expensive or impractical.

%\paragraph{Model Size Comparison}
Model scale is a key driver of performance, particularly within the same model family.
Larger models have been shown to more effectively memorize and generalize low-resource language phenomena, consistent with neural scaling laws \citep{gordon2021data, kaplan2020scaling}.
However, scale alone does not guarantee superior results. Our findings not only indicate some performance saturation among larger models (above 30B), but also highlight the potential of thinking-enabled models, where even relatively small Qwen3 variants capture sufficient knowledge to rival or surpass much larger counterparts, thanks to explicit reasoning traces. Besides the thinking mode, architectural and multimodal training strategies play a central role. For instance, Qwen3-A22B and A3B exploit a Mixture-of-Experts (MoE) design to combine high reasoning efficiency with strong accuracy. Likewise, when comparing Qwen2.5-VL-7B against InternVL3-8B, which shares the same language model backbone, the Qwen variant proves more robust on visual signals. %These results suggest that reasoning mechanisms and targeted training objectives, rather than raw parameter counts, now define performance boundaries.
% todo: recheck all the facts

% \paragraph{Modality Comparison}
Multimodal inputs (I+T) generally improve performance, but outcomes vary with models. In the InternVL3 series and Gemma3, performance with image-only input lags far behind multimodal input for the same models, reflecting a limited ability to derive the required information solely from images. A different, even worse issue arises in Gemma3, where multimodal fusion degrades performance compared to text-only, likely due to overreliance on noisy visual features. By contrast, Qwen2.5-VL shows superior integration, successfully featuring a visual-only capability powerful enough to match text-only performance, and complementing text robustness with accurate visual cues. These mixed results underline both the difficulty of historical OCR and image-based Latin extraction, and the importance of our dataset for testing multimodal performance with preprocessed OCR text input.

Finally, we address the issue of robustness in rejecting non-Latin pages. Beyond standard detection metrics, the \textbf{NL. F1} column reveals a divergence in model behavior: while detection recall is generally high, the ability to correctly reject non-Latin pages varies. This instability is most pronounced in Gemma3 with image-only input. It corroborates our earlier observation regarding its weak visual grounding. Yet, this issue extends beyond a single model, and we analyze further in Section \ref{sec:non-latin}.

\subsection{Behavior on Non-Latin Pages}
\label{sec:non-latin}

To further quantify the over-sensitivity observed in the main results, we analyzed non-Latin class recall (for page-level detection) and false positive token rates (for false Latin segment extraction) as two additional metrics, shown in Table~\ref{tab:additional_metrics}. The analysis reveals that most models tend to over-detect Latin, particularly the smaller ones. The statistical baseline Lingua performs poorly, misclassifying over 70\% of these pages, which would be a significant issue for large-scale processing.

More critically, multimodal inputs can overwhelm smaller models: Qwen2.5-VL-7B  suffers a catastrophic collapse, indicating that overlapped visual and textual signals cause the model to hallucinate Latin across half the page. Crucially, this is a failure of fusion rather than perception: when fed with only images, the same model recovers much of its rejection capability, confirming that long and noisy OCR text acts as a distractor that overrides visual grounding in smaller architectures.

In contrast, the most robust performance comes from Qwen2.5-32B with text-only input, likely due to a stronger sensitivity to linguistic context provided by model scale. However, a promising finding is that even on misclassified pages, the number of erroneously extracted Latin tokens by the LLMs is generally small, suggesting that simple downstream filtering could effectively mitigate this over-detection problem.

\subsection{Performance by Category}
\label{sec:cat_res}
%\vspace{8pt}

Shown as Figure~\ref{fig:cat_res}, we evaluate model performance across different functional text categories specified in Section~\ref{sec:categories}. As the models are tasked with extraction rather than classification, we measure performance using token-level recall in categorized segments on each page by the same fuzzy matching process as stated in Section~\ref{sec:metrics}. 
There is a large disparity between category difficulty: the models yield almost perfect performance for longer text types like independent and bilingual categories while struggling with the typically shorter code-switching, dictionaries, tables and charts, side-notes, and to a lesser degree with footnotes. 

The consistent performance trend across all models, including the baseline, suggests a shared, statistically driven behavior that relies on Latin's vocabulary and paragraph patterns over deep functional semantic understanding. Under this view, vision's role is mainly to improve OCR accuracy, and the explicit thinking only helps invoke more basic knowledge of Latin. To probe this hypothesis, we designed a more demanding joint extraction and categorization task (see Appendix~\ref{appendix:cat_res} for the details). The resulting categorized token F1 scores were exceptionally low: 21.0\% for DeepSeek-R1 and 14.6\% for Qwen2.5-VL (I+T), which strongly support our hypothesis. We conclude that this weak functional understanding, compounded with known OCR challenges, is a key factor behind the low extraction recall in certain categories. This conclusion is also partly validated by our prompt engineering experiments in Section~\ref{sec:prompts}.

\begin{table}[t!] % updated
\centering
\begin{adjustbox}{max width=0.45\textwidth}
\begin{tabular}{l|l|l|c|c|c}
\toprule
\multicolumn{4}{c|}{\textsc{Model Details}} & \textsc{Page} & \textsc{Token} \\
\hline
\textbf{Model} & \textbf{Variant} & \textbf{Size} & \textbf{Mode} & \textbf{NL. Recall $\uparrow$} &  \textbf{FP Rate $\downarrow$} \\
\midrule
Lingua & - & - & T & 27.69 & 2.64\\
\midrule
DeepSeek-R1 & - & 671B & T & 77.69 & 0.43 \\
\midrule
Qwen3 & - & 4B & T & 40.77 & 4.97 \\
\midrule
\multirow{6}{*}{Qwen2.5} 
 & VL & 32B & I+T & 65.38 & 2.66 \\
 & VL & 32B & I & 73.08 & 2.09 \\
 & - & 32B & T & \textbf{96.15} & \textbf{0.17} \\
\cmidrule(lr){2-6}
& VL & 7B & I+T & 10.77 & 55.54\\
& VL & 7B & I & 85.38 &	2.97 \\
& - & 7B & T & 66.15 & 10.07 \\
\bottomrule
\end{tabular}
\end{adjustbox}
\caption{Analysis on non-Latin pages: \textbf{Non-Latin Recall} in pages and \textbf{False Positive Rate} showing averaged percentage of tokens falsely identified as Latin on truly non‑Latin pages.}
\label{tab:additional_metrics}
\vspace{-1em}
\end{table} 

\subsection{Impact of Prompt Variations}
\label{sec:prompts}

To assess the robustness of our findings, we first evaluated the performance stability of four representative models across six distinct prompt strategies, as summarized in Table~\ref{tab:prompt_variance}. We observe a consistent insensitivity to instruction phrasing across diverse architectures, with minimal Token F1 variance ($\sigma < 1.1$) and no consistent gains, for both text-only and multimodal models. This statistical evidence suggests that extraction capabilities are largely anchored by the models' intrinsic domain knowledge rather than specific prompt engineering.

\begin{figure}[t!]
    \centering
    \includegraphics[width=0.48\textwidth]{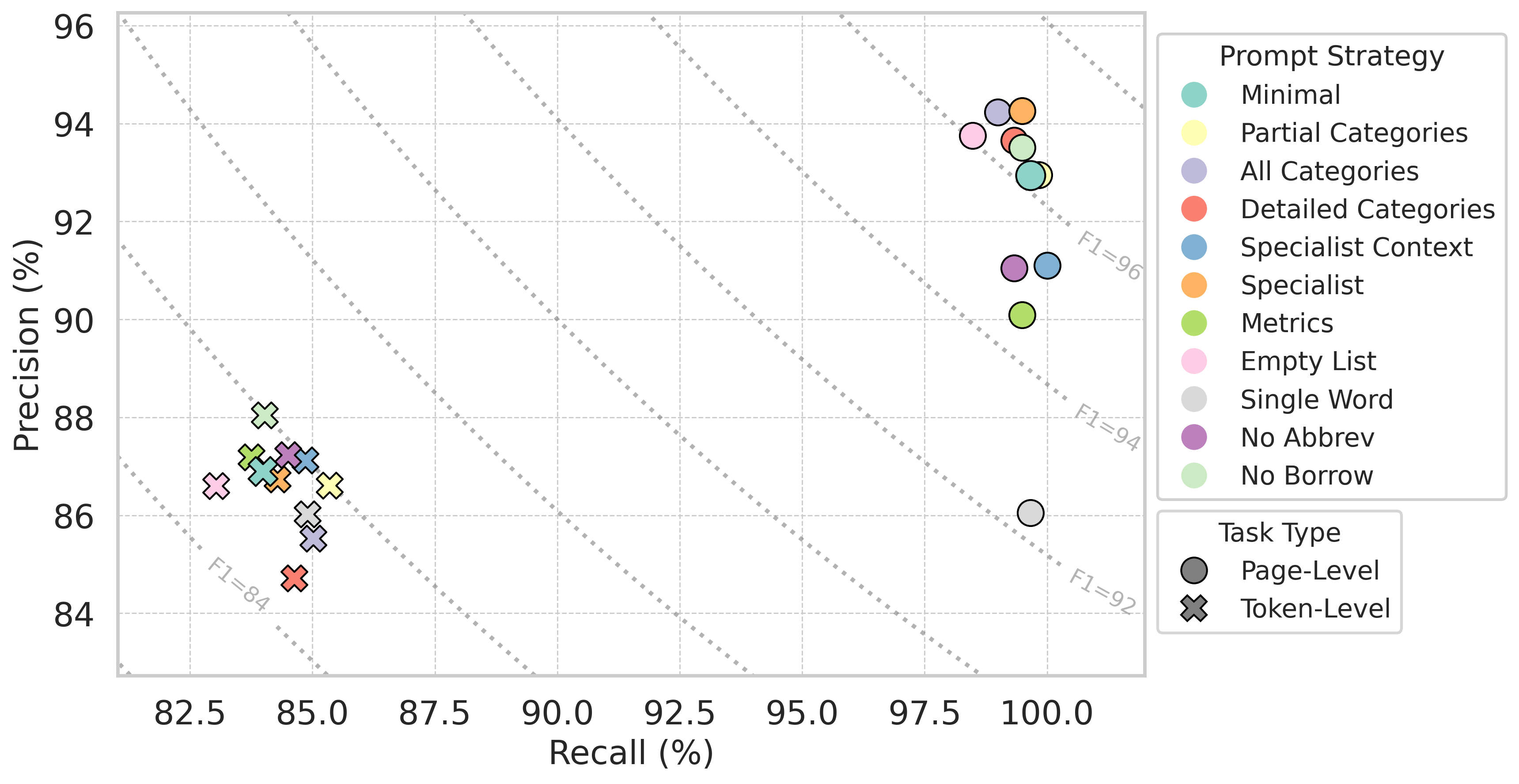}
    \caption{Impact of prompting on Qwen2.5-VL-32B.} 
    \label{fig:prompts}
    \vspace{-1em}
\end{figure}

Focusing on the detailed behavior of the top-performing Qwen2.5-VL-32B model with multimodal input (shown in Figure~\ref{fig:prompts}), most strategies result in only marginal changes, typically a trade-off between precision and recall, or different tasks, rather than a definitive improvement. We highlight the most salient strategies here, with full tables of all the prompts and results available in Appendix~\ref{appendix:prompt_exp}. For instance, instructing the model to include ``single-word'' segments lowered page-level precision by encouraging over-extraction on short segments, while negative constraints like ``no borrow'' slightly increased token-level precision through a more conservative behavior by excluding the borrowed Latin words in other languages. Most revealingly, the ``partial categories'' strategy, directing the model to focus on its known weak areas, increased token recall but at the cost of a drop in precision. This indicates the prompt did not grant the model a deeper functional understanding; instead, it likely encouraged more speculative guessing in targeted contexts. This outcome reinforces our central hypothesis stated in Section~\ref{sec:cat_res} that the model's extraction is guided by statistical patterns rather than a semantic comprehension of the text's function.

\begin{table}[t!]
\centering
\small
\begin{adjustbox}{max width=0.45\textwidth}
\begin{tabular}{l|c|c|c||c|c}
\toprule
\textbf{Model} & \textbf{Variant} & \textbf{Size} & \textbf{Mode} & \textbf{Page F1} & \textbf{Tok. F1} \\
\midrule
DeepSeek-R1 & Distill & 70B & T & $96.28_{\pm 1.06}$ & $84.27_{\pm 0.20}$ \\
Qwen3 & - & 14B & T & $96.47_{\pm 1.00}$ & $84.42_{\pm 0.62}$ \\
Qwen2.5 & - & 32B & T & $97.14_{\pm 0.75}$ & $81.89_{\pm 1.06}$ \\
\midrule
\multirow{2}{*}{Qwen2.5} & VL & 32B & I+T & $95.58_{\pm 1.46}$ & $84.44_{\pm 0.55}$ \\
 & VL & 32B & I & $95.72_{\pm 1.79}$ & $79.51_{\pm 0.79}$ \\
\bottomrule
\end{tabular}
\end{adjustbox}
\caption{Prompt robustness statistics: performance reported as $Mean_{\pm Std}$ across 6 prompt strategies.}
\vspace{-1em}
\label{tab:prompt_variance}
\end{table}

\subsection{Impact of OCR Quality}
\label{sec:ocr_impact}

As detailed in Section~\ref{sec:ocr_setting}, evaluating models on the ECCO corpus is complicated by significant discrepancies between legacy OCR and modern visual processing. To ensure meaningful assessment across all model types, we employ an OpenAI o1-based post-correction pipeline~\cite{kanerva2025ocr}.

To quantify the necessity of this normalization, we conducted an ablation study using raw, uncorrected OCR transcripts, shown as Table~\ref{tab:ocr_ablation_f1}. We observe that raw OCR creates a noise barrier that affects all models utilizing text input (including Vision-Language models fed with OCR text), with pure text models suffering the most severe degradation ($>14$ point drop). Consequently, the performance degrades so severely that text-dependent models fall behind the best image-only model. This confirms that raw ECCO OCR constructs an excessive noise barrier that surpasses the detection task itself. Thus, our post-correction pipeline is not an o1-assisted ``shortcut'' for OCR-dependent models, but a necessary normalization to ensure fair cross-modal comparison. 

\begin{table}[t]
\centering
\small
\setlength{\tabcolsep}{3pt}
\begin{adjustbox}{max width=0.45\textwidth}
\begin{tabular}{lccc} % todo: check the numbers
\toprule
\textbf{Model \& Input} & \textbf{Page F1} & \textbf{Tok. F1} & \textbf{$\Delta$ Tok. F1} \\
\midrule
\multicolumn{4}{l}{\textit{Text-only Models}} \\
\textbf{DeepSeek-R1} & 96.65 & 72.26 & \multirow{2}{*}{\textcolor{red}{-14.74}} \\
\quad w/ Cleaned OCR & 97.53 & 87.00 & \\
\textbf{Qwen3-14B} & 93.33 & 67.12 & \multirow{2}{*}{\textcolor{red}{-17.66}} \\
\quad w/ Cleaned OCR & 96.61 & 84.78 & \\
\textbf{Qwen2.5-32B} & 96.02 & 67.48 & \multirow{2}{*}{\textcolor{red}{-14.09}} \\
\quad w/ Cleaned OCR & 97.26 & 81.57 & \\
\midrule
\multicolumn{4}{l}{\textit{Vision-Language Models}} \\
\textbf{Qwen2.5-VL-32B} & 95.38 & 79.24 & \multirow{2}{*}{\textcolor{red}{-5.08}} \\
\quad w/ Cleaned OCR & 96.18 & 84.32 & \\
\textbf{InternVL3-38B} & 93.74 & 74.56 & \multirow{2}{*}{\textcolor{red}{-9.68}} \\
\quad w/ Cleaned OCR & 95.62 & 84.24 & \\
\midrule
\multicolumn{4}{l}{\textit{Image-only Models}} \\
\textbf{Qwen2.5-VL-32B (I)} & \textbf{96.80} & \textbf{79.82} & - \\
\textbf{InternVL3-38B (I)} & 89.51 & 56.53 & - \\
\bottomrule
\end{tabular}
\end{adjustbox}
\caption{Ablation on OCR quality. $\Delta$ Tok. F1 denotes the token-level F1 change compared to the models with cleaned OCR text input.}
\vspace{-1em}
\label{tab:ocr_ablation_f1}
\end{table}

\subsection{Qualitative Evaluation and Error Analysis}

Our qualitative evaluation reveals that LLMs' performance is primarily limited by two interacting factors: the inherent challenges of the ECCO data and the models' systematic misinterpretations of the task. These issues set practical limits on upper-bound scores. See Appendix~\ref{appendix:qualitative} for a detailed, example-oriented discussion.

First, data-centric challenges stem from the ECCO collection. Poor image quality and complex page layouts, such as multi-column text, marginalia, and varied fonts (see Figures~\ref{fig:example-page2} and~\ref{fig:example-page3} in Appendix~\ref{appendix:append_data}), result in noisy and fragmented OCR, even after post-correction. This degradation directly impacts model performance and hampers reliable annotation, especially for the brief Latin snippets found in challenging categories like dictionaries or footnotes, which are more susceptible to severe OCR errors (see Figure~\ref{fig:example-page4}).

Second, we observe model-centric challenges. Models appear to rely on vocabulary and paragraph patterns over a deep functional understanding, leading to poor performance on short fragments and on words English loaned from Latin. This results in a consistent definitional mismatch with our annotation guidelines: models frequently misidentify Roman named entities, common anglicized Latin phrases (e.g., ``e.g.'', ``etc.'', etc.), and Latin-derived jargon as Latin, which significantly harms precision (see Figure~\ref{fig:example-page5}). In many instances, this definitional disagreement accounts for the entirety of the prediction error, suggesting that model precision could be theoretically stronger with minor adjustments to the task definition in prompting, a behavior also empirically observed during our prompt tuning experiments, e.g., ``no borrow'' strategy (Section~\ref{sec:prompts}).

\section{Conclusion}
This paper introduced and benchmarked the novel task of zero-shot Latin discovery in historical documents. Our analysis demonstrates that this task is solvable with excellent performance of LLMs without task-specific fine-tuning. However, our key finding is that this success appears to stem not from deep functional semantic understanding, but from the models' ability to leverage more superficial statistical cues like vocabulary and text patterns, a behavior similar to traditional methods. This may limit the current models' approach to more nuanced historical tasks, suggesting that foundation models cannot replace human interpretative expertise.

A particular implication of our work is the high performance of image-only models (up to approximately 99\% page-level F1), which enables an efficient two-stage approach for processing vast non-OCR archives. This capability may effectively unlock the ``dark archives'' of digitized heritage. Building on these, our future work will focus on applying our pipeline to the entire ECCO collection and extending the methodology to other corpora, such as the French BnF's collection. 

% This paper presented a novel task and a dataset for Latin extraction from early-modern book pages. We systematically evaluated diverse foundational models and found that this task can be solved with excellent performance, without fine-tuning. However, such performance can be achieved only with bigger models (32B parameters).

% Our results show that 94\% F1 performance on page-level detection can be achieved with only image input. This has practical implications since a visual model can be used even in collections that have no OCR, or whose OCR is not of sufficient quality. In this case, OCR or post-OCR correction can be performed only on pages preselected by a visual model, which would save computation cost.

% We also found that semantic analysis is crucial to distinguish actual Latin phrases from usages of Latin words in English text. 
% %
% Thus we argue for the usage of large foundation models, or a pipeline of visual and textual models, rather than pure statistical methods.

% Further steps will include processing the whole ECCO collection and publishing a complete dataset of Latin fragments in the ECCO books. We also plan to expand our results to other Early-modern collections, such as the French BNF collection.

\section*{Limitations}
% required section, does not contribute to the page limit

Our findings rely on a single corpus (18th-century British ECCO). While this aligns with our research scope, corpus-specific biases may limit generalizability, necessitating future validation on diverse collections, such as Romance-language texts. Additionally, due to the high cost of annotation, we lacked a separate validation set for hyperparameter tuning. Model selection relied on literature baselines rather than independent optimization.

Computational constraints limited us to single experimental runs with deterministic settings. While ensuring reproducibility, this approach does not capture the potential variance inherent in stochastic generation. Finally, this study benchmarks off-the-shelf models in a zero-shot setting. We acknowledge that task-specific fine-tuning would likely yield higher performance ceilings. Moreover, such experiments would be invaluable for identifying the most persistent challenges of the task that remain even after direct, task-specific training—perhaps related to deep functional semantic understanding of Latin—thereby providing crucial insights for future model development. Our newly created dataset provides the first dedicated resource to enable this line of inquiry.

\section*{Ethical Considerations}

The underlying literary works from which our dataset is derived, sourced from 18th-century texts within the Eighteenth Century Collections Online (ECCO), are in the public domain. The compilation and sharing of our dataset, which comprises annotated excerpts and portions of page images from this collection, are conducted for research purposes under the permissions granted. We are committed to ensuring that the creation and dissemination of this dataset adhere to relevant copyright considerations and ethical guidelines.

We used ChatGPT and Gemini for grammar and spell-checking and stylistic polishing of the draft of this manuscript. All suggestions were critically reviewed and edited by the authors to ensure factual accuracy and originality.

\section*{Acknowledgments}
We thank the Area Chair and the anonymous reviewers for their constructive comments and suggestions. This project has received funding from the European Union's Horizon Europe programme for research and innovation under MSCA Doctoral Networks 2022, Grant Agreement No. 101120349 and Grant Agreement No. 101119511. We also acknowledge CSC – IT Center for Science, Finland, for awarding this project access to the LUMI supercomputer, owned by the EuroHPC Joint Undertaking, hosted by CSC (Finland) and the LUMI consortium.

\bibliography{citation,anthology}

\begin{thebibliography}{41}
\providecommand{\natexlab}[1]{#1}

\bibitem[{Achiam et~al.(2023)Achiam, Adler, Agarwal, Ahmad, Akkaya, Aleman, Almeida, Altenschmidt, Altman, Anadkat et~al.}]{achiam2023gpt}
Josh Achiam, Steven Adler, Sandhini Agarwal, Lama Ahmad, Ilge Akkaya, Florencia~Leoni Aleman, Diogo Almeida, Janko Altenschmidt, Sam Altman, Shyamal Anadkat, et~al. 2023.
\newblock {GPT-4} technical report.
\newblock \emph{arXiv preprint arXiv:2303.08774}.

\bibitem[{Aguilar et~al.(2020)Aguilar, Kar, and Solorio}]{aguilar-etal-2020-lince}
Gustavo Aguilar, Sudipta Kar, and Thamar Solorio. 2020.
\newblock \href {https://aclanthology.org/2020.lrec-1.223/} {{L}in{CE}: A centralized benchmark for linguistic code-switching evaluation}.
\newblock In \emph{Proceedings of the Twelfth Language Resources and Evaluation Conference}, pages 1803--1813, Marseille, France. European Language Resources Association.

\bibitem[{Backer and Hyman(2025)}]{backer2025bootstrapping}
Samuel Backer and Louis Hyman. 2025.
\newblock Bootstrapping {AI}: Interdisciplinary approaches to assessing {OCR} quality in english-language historical documents.
\newblock In \emph{Proceedings of the 5th International Conference on Natural Language Processing for Digital Humanities}, pages 251--256.

\bibitem[{Bai et~al.(2025)Bai, Chen, Liu, Wang, Ge, Song, Dang, Wang, Wang, Tang, Zhong, Zhu, Yang, Li, Wan, Wang, Ding, Fu, Xu, Ye, Zhang, Xie, Cheng, Zhang, Yang, Xu, and Lin}]{Qwen2.5-VL}
Shuai Bai, Keqin Chen, Xuejing Liu, Jialin Wang, Wenbin Ge, Sibo Song, Kai Dang, Peng Wang, Shijie Wang, Jun Tang, Humen Zhong, Yuanzhi Zhu, Mingkun Yang, Zhaohai Li, Jianqiang Wan, Pengfei Wang, Wei Ding, Zheren Fu, Yiheng Xu, Jiabo Ye, Xi~Zhang, Tianbao Xie, Zesen Cheng, Hang Zhang, Zhibo Yang, Haiyang Xu, and Junyang Lin. 2025.
\newblock {Qwen2.5-VL} technical report.
\newblock \emph{arXiv preprint arXiv:2502.13923}.

\bibitem[{Barman et~al.(2014)Barman, Das, Wagner, and Foster}]{barman-etal-2014-code}
Utsab Barman, Amitava Das, Joachim Wagner, and Jennifer Foster. 2014.
\newblock \href {https://doi.org/10.3115/v1/W14-3902} {Code mixing: A challenge for language identification in the language of social media}.
\newblock In \emph{Proceedings of the First Workshop on Computational Approaches to Code Switching}, pages 13--23, Doha, Qatar. Association for Computational Linguistics.

\bibitem[{Boros et~al.(2024)Boros, Ehrmann, Romanello, Najem-Meyer, and Kaplan}]{boros-etal-2024-post}
Emanuela Boros, Maud Ehrmann, Matteo Romanello, Sven Najem-Meyer, and Fr{\'e}d{\'e}ric Kaplan. 2024.
\newblock \href {https://aclanthology.org/2024.latechclfl-1.14/} {Post-correction of historical text transcripts with large language models: An exploratory study}.
\newblock In \emph{Proceedings of the 8th Joint SIGHUM Workshop on Computational Linguistics for Cultural Heritage, Social Sciences, Humanities and Literature (LaTeCH-CLfL 2024)}, pages 133--159, St. Julians, Malta. Association for Computational Linguistics.

\bibitem[{Burns(2023)}]{burns2023latincy}
Patrick~J Burns. 2023.
\newblock Latincy: Synthetic trained pipelines for {Latin NLP}.
\newblock \emph{arXiv preprint arXiv:2305.04365}.

\bibitem[{Burns et~al.(2021)Burns, Brofos, Li, Chaudhuri, and Dexter}]{burns2021profiling}
Patrick~J Burns, James~A Brofos, Kyle Li, Pramit Chaudhuri, and Joseph~P Dexter. 2021.
\newblock Profiling of intertextuality in {Latin} literature using word embeddings.
\newblock In \emph{Proceedings of the 2021 Conference of the North American Chapter of the Association for Computational Linguistics: Human Language Technologies}, pages 4900--4907.

\bibitem[{Contributors(2023)}]{2023opencompass}
OpenCompass Contributors. 2023.
\newblock {OpenCompass}: A universal evaluation platform for foundation models.
\newblock \url{https://github.com/open-compass/opencompass}.

\bibitem[{Gordon et~al.(2021)Gordon, Duh, and Kaplan}]{gordon2021data}
Mitchell~A Gordon, Kevin Duh, and Jared Kaplan. 2021.
\newblock Data and parameter scaling laws for neural machine translation.
\newblock In \emph{Proceedings of the 2021 Conference on Empirical Methods in Natural Language Processing}, pages 5915--5922.

\bibitem[{Gorovaia et~al.(2024)Gorovaia, Schmidt, and Yamshchikov}]{gorovaia-etal-2024-sui}
Svetlana Gorovaia, Gleb Schmidt, and Ivan~P. Yamshchikov. 2024.
\newblock \href {https://doi.org/10.18653/v1/2024.nlp4dh-1.39} {{S}ui generis: Large language models for authorship attribution and verification in {L}atin}.
\newblock In \emph{Proceedings of the 4th International Conference on Natural Language Processing for Digital Humanities}, pages 398--412, Miami, USA. Association for Computational Linguistics.

\bibitem[{Guo et~al.(2025)Guo, Yang, Zhang, Song, Wang, Zhu, Xu, Zhang, Ma, Bi et~al.}]{guo2025deepseek}
Daya Guo, Dejian Yang, Haowei Zhang, Junxiao Song, Peiyi Wang, Qihao Zhu, Runxin Xu, Ruoyu Zhang, Shirong Ma, Xiao Bi, et~al. 2025.
\newblock {DeepSeek-R1} incentivizes reasoning in {LLM}s through reinforcement learning.
\newblock \emph{Nature}, 645(8081):633--638.

\bibitem[{Guzm{\'a}n et~al.(2017)Guzm{\'a}n, Ricard, Serigos, Bullock, and Toribio}]{guzman2017metrics}
Gualberto~A Guzm{\'a}n, Joseph Ricard, Jacqueline Serigos, Barbara~E Bullock, and Almeida~Jacqueline Toribio. 2017.
\newblock Metrics for modeling code-switching across corpora.
\newblock In \emph{Interspeech}, pages 67--71.

\bibitem[{Hauser et~al.(2024)Hauser, Kondor, Reddish, Benam, Cioni, Villa, Bennett, Hoyer, Francois, Turchin et~al.}]{hauser2024large}
Jakob Hauser, Daniel Kondor, Jenny Reddish, Majid Benam, Enrico Cioni, Federica Villa, James Bennett, Daniel Hoyer, Pieter Francois, Peter Turchin, et~al. 2024.
\newblock Large language models' expert-level global history knowledge benchmark {(HiST-LLM)}.
\newblock \emph{Advances in Neural Information Processing Systems}, 37:32336--32369.

\bibitem[{Jaech et~al.(2024)Jaech, Kalai, Lerer, Richardson, El-Kishky, Low, Helyar, Madry, Beutel, Carney et~al.}]{jaech2024openai}
Aaron Jaech, Adam Kalai, Adam Lerer, Adam Richardson, Ahmed El-Kishky, Aiden Low, Alec Helyar, Aleksander Madry, Alex Beutel, Alex Carney, et~al. 2024.
\newblock {OpenAI o1} system card.
\newblock \emph{arXiv preprint arXiv:2412.16720}.

\bibitem[{Johnson et~al.(2021)Johnson, Burns, Stewart, Cook, Besnier, and Mattingly}]{johnson-etal-2021-classical}
Kyle~P. Johnson, Patrick~J. Burns, John Stewart, Todd Cook, Cl{\'e}ment Besnier, and William J.~B. Mattingly. 2021.
\newblock \href {https://doi.org/10.18653/v1/2021.acl-demo.3} {The {C}lassical {L}anguage {T}oolkit: {A}n {NLP} framework for pre-modern languages}.
\newblock In \emph{Proceedings of the 59th Annual Meeting of the Association for Computational Linguistics and the 11th International Joint Conference on Natural Language Processing: System Demonstrations}, pages 20--29, Online. Association for Computational Linguistics.

\bibitem[{Kanerva et~al.(2025)Kanerva, Ledins, K{\"a}pyaho, and Ginter}]{kanerva2025ocr}
Jenna Kanerva, Cassandra Ledins, Siiri K{\"a}pyaho, and Filip Ginter. 2025.
\newblock {OCR} error post-correction with {LLM}s in historical documents: No free lunches.
\newblock \emph{arXiv preprint arXiv:2502.01205}.

\bibitem[{Kaplan et~al.(2020)Kaplan, McCandlish, Henighan, Brown, Chess, Child, Gray, Radford, Wu, and Amodei}]{kaplan2020scaling}
Jared Kaplan, Sam McCandlish, Tom Henighan, Tom~B Brown, Benjamin Chess, Rewon Child, Scott Gray, Alec Radford, Jeffrey Wu, and Dario Amodei. 2020.
\newblock Scaling laws for neural language models.
\newblock \emph{arXiv preprint arXiv:2001.08361}.

\bibitem[{Kupari et~al.(2024)Kupari, Henriksson, Laippala, and Kanerva}]{kupari-etal-2024-improving}
Hanna-Mari~Kristiina Kupari, Erik Henriksson, Veronika Laippala, and Jenna Kanerva. 2024.
\newblock \href {https://doi.org/10.18653/v1/2024.nlp4dh-1.21} {Improving {L}atin dependency parsing by combining treebanks and predictions}.
\newblock In \emph{Proceedings of the 4th International Conference on Natural Language Processing for Digital Humanities}, pages 216--228, Miami, USA. Association for Computational Linguistics.

\bibitem[{Kwon et~al.(2023)Kwon, Li, Zhuang, Sheng, Zheng, Yu, Gonzalez, Zhang, and Stoica}]{kwon2023efficient}
Woosuk Kwon, Zhuohan Li, Siyuan Zhuang, Ying Sheng, Lianmin Zheng, Cody~Hao Yu, Joseph~E. Gonzalez, Hao Zhang, and Ion Stoica. 2023.
\newblock Efficient memory management for large language model serving with {PagedAttention}.
\newblock In \emph{Proceedings of the ACM SIGOPS 29th Symposium on Operating Systems Principles}.

\bibitem[{Luo et~al.(2024)Luo, Shen, Zhu, Zheng, Yu, and Yao}]{luo2024layoutllm}
Chuwei Luo, Yufan Shen, Zhaoqing Zhu, Qi~Zheng, Zhi Yu, and Cong Yao. 2024.
\newblock {LayoutLLM}: Layout instruction tuning with large language models for document understanding.
\newblock In \emph{Proceedings of the IEEE/CVF conference on computer vision and pattern recognition}, pages 15630--15640.

\bibitem[{Marjanen et~al.(2025)Marjanen, Tahko, Lahti, and Tolonen}]{marjanen2025book}
Jani Marjanen, Tuuli Tahko, Leo Lahti, and Mikko Tolonen. 2025.
\newblock Book printing in {Latin} and vernacular languages in northern {Europe}, 1500--1800.
\newblock In \emph{The Hermeneutics of Bibliographic Data and Cultural Metadata}, pages 27--66. National Library of Norway.

\bibitem[{Perrone et~al.(2021)Perrone, Hengchen, Palma, Vatri, Smith, and McGillivray}]{perrone2021lexical}
Valerio Perrone, Simon Hengchen, Marco Palma, Alessandro Vatri, Jim~Q Smith, and Barbara McGillivray. 2021.
\newblock Lexical semantic change for {Ancient Greek} and {Latin}.
\newblock \emph{Computational approaches to semantic change}, 6.

\bibitem[{Rosson et~al.(2023)Rosson, M{\"a}kel{\"a}, Vaara, Mahadevan, Ryan, and Tolonen}]{rosson2023reception}
David Rosson, Eetu M{\"a}kel{\"a}, Ville Vaara, Ananth Mahadevan, Yann Ryan, and Mikko Tolonen. 2023.
\newblock Reception reader: Exploring text reuse in early modern {British} publications.
\newblock \emph{Journal of Open Humanities Data}, 9(1).

\bibitem[{Schulz and Keller(2016)}]{schulz-keller-2016-code}
Sarah Schulz and Mareike Keller. 2016.
\newblock \href {https://doi.org/10.18653/v1/W16-2105} {Code-switching ubique est - language identification and part-of-speech tagging for historical mixed text}.
\newblock In \emph{Proceedings of the 10th {SIGHUM} Workshop on Language Technology for Cultural Heritage, Social Sciences, and Humanities}, pages 43--51, Berlin, Germany. Association for Computational Linguistics.

\bibitem[{Sprugnoli et~al.(2024)Sprugnoli, Iurescia, and Passarotti}]{sprugnoli-etal-2024-overview}
Rachele Sprugnoli, Federica Iurescia, and Marco Passarotti. 2024.
\newblock \href {https://aclanthology.org/2024.lt4hala-1.21/} {Overview of the {E}va{L}atin 2024 evaluation campaign}.
\newblock In \emph{Proceedings of the Third Workshop on Language Technologies for Historical and Ancient Languages (LT4HALA) @ LREC-COLING-2024}, pages 190--197, Torino, Italia. ELRA and ICCL.

\bibitem[{Stahl(2021)}]{pemistahl2021lingua}
Peter~M. Stahl. 2021.
\newblock {Lingua}: The most accurate natural language detection library for {Python}.
\newblock \url{https://github.com/pemistahl/lingua-py}.
\newblock Python bindings for the Lingua language detection library.

\bibitem[{Sterner and Teufel(2023)}]{sterner-teufel-2023-tongueswitcher}
Igor Sterner and Simone Teufel. 2023.
\newblock \href {https://aclanthology.org/2023.calcs-1.1/} {{T}ongue{S}witcher: Fine-grained identification of {G}erman-{E}nglish code-switching}.
\newblock In \emph{Proceedings of the 6th Workshop on Computational Approaches to Linguistic Code-Switching}, pages 1--13, Singapore. Association for Computational Linguistics.

\bibitem[{Straka and Strakov{\'a}(2020)}]{straka-strakova-2020-udpipe}
Milan Straka and Jana Strakov{\'a}. 2020.
\newblock \href {https://aclanthology.org/2020.lt4hala-1.20/} {{UDP}ipe at {E}va{L}atin 2020: Contextualized embeddings and treebank embeddings}.
\newblock In \emph{Proceedings of LT4HALA 2020 - 1st Workshop on Language Technologies for Historical and Ancient Languages}, pages 124--129, Marseille, France. European Language Resources Association (ELRA).

\bibitem[{St{\"u}ssi and Str{\"o}bel(2024)}]{stussi-strobel-2024-part}
Elina St{\"u}ssi and Phillip Str{\"o}bel. 2024.
\newblock \href {https://aclanthology.org/2024.latechclfl-1.18/} {Part-of-speech tagging of 16th-century {L}atin with {GPT}}.
\newblock In \emph{Proceedings of the 8th Joint SIGHUM Workshop on Computational Linguistics for Cultural Heritage, Social Sciences, Humanities and Literature (LaTeCH-CLfL 2024)}, pages 196--206, St. Julians, Malta. Association for Computational Linguistics.

\bibitem[{Team et~al.(2025)Team, Kamath, Ferret, Pathak, Vieillard, Merhej, Perrin, Matejovicova, Ram{\'e}, Rivi{\`e}re et~al.}]{team2025gemma}
Gemma Team, Aishwarya Kamath, Johan Ferret, Shreya Pathak, Nino Vieillard, Ramona Merhej, Sarah Perrin, Tatiana Matejovicova, Alexandre Ram{\'e}, Morgane Rivi{\`e}re, et~al. 2025.
\newblock Gemma 3 technical report.
\newblock \emph{arXiv preprint arXiv:2503.19786}.

\bibitem[{Team(2025)}]{qwen3technicalreport}
Qwen Team. 2025.
\newblock \href {https://arxiv.org/abs/2505.09388} {Qwen3 technical report}.
\newblock \emph{Preprint}, arXiv:2505.09388.

\bibitem[{Tito et~al.(2021)Tito, Karatzas, and Valveny}]{tito2021document}
Rub{\`e}n Tito, Dimosthenis Karatzas, and Ernest Valveny. 2021.
\newblock Document collection visual question answering.
\newblock In \emph{International Conference on Document Analysis and Recognition}, pages 778--792. Springer.

\bibitem[{Tkachenko et~al.(2020-2025)Tkachenko, Malyuk, Holmanyuk, and Liubimov}]{LabelStudio}
Maxim Tkachenko, Mikhail Malyuk, Andrey Holmanyuk, and Nikolai Liubimov. 2020-2025.
\newblock \href {https://github.com/HumanSignal/label-studio} {{Label Studio}: Data labeling software}.
\newblock Open source software available from https://github.com/HumanSignal/label-studio.

\bibitem[{Tolonen et~al.(2022)Tolonen, M{\"a}kel{\"a}, and Lahti}]{tolonen2022anatomy}
Mikko Tolonen, Eetu M{\"a}kel{\"a}, and Leo Lahti. 2022.
\newblock The anatomy of {Eighteenth Century Collections Online (ECCO)}.
\newblock \emph{Eighteenth-century studies}, 56(1):95--123.

\bibitem[{Volk et~al.(2024)Volk, Fischer, Fischer, Scheurer, and Str{\"o}bel}]{volk-etal-2024-llm}
Martin Volk, Dominic~Philipp Fischer, Lukas Fischer, Patricia Scheurer, and Phillip~Benjamin Str{\"o}bel. 2024.
\newblock \href {https://aclanthology.org/2024.lt4hala-1.15/} {{LLM}-based machine translation and summarization for {L}atin}.
\newblock In \emph{Proceedings of the Third Workshop on Language Technologies for Historical and Ancient Languages (LT4HALA) @ LREC-COLING-2024}, pages 122--128, Torino, Italia. ELRA and ICCL.

\bibitem[{Volk et~al.(2022)Volk, Fischer, Scheurer, Schroffenegger, Schwitter, Str{\"o}bel, and Suter}]{volk-etal-2022-nunc}
Martin Volk, Lukas Fischer, Patricia Scheurer, Bernard~Silvan Schroffenegger, Raphael Schwitter, Phillip Str{\"o}bel, and Benjamin Suter. 2022.
\newblock \href {https://aclanthology.org/2022.lrec-1.311/} {Nunc profana tractemus. detecting code-switching in a large corpus of 16th century letters}.
\newblock In \emph{Proceedings of the Thirteenth Language Resources and Evaluation Conference}, pages 2901--2908, Marseille, France. European Language Resources Association.

\bibitem[{Xie et~al.(2025)Xie, La~Mela, and Tell}]{xie2025multimodal}
Yunting Xie, Matti La~Mela, and Fredrik Tell. 2025.
\newblock Multimodal {LLM}-assisted information extraction from historical documents: The case of {Swedish} patent cards (1945-1975) and {ChatGPT}.
\newblock In \emph{The 9th Digital Humanities in the Nordic and Baltic Countries Conference (DHNB 2025), March 5--7, 2025, Tartu, Estonia}, pages 1--15. University of Oslo Library.

\bibitem[{Yue et~al.(2024)Yue, Ni, Zhang, Zheng, Liu, Zhang, Stevens, Jiang, Ren, Sun et~al.}]{yue2024mmmu}
Xiang Yue, Yuansheng Ni, Kai Zhang, Tianyu Zheng, Ruoqi Liu, Ge~Zhang, Samuel Stevens, Dongfu Jiang, Weiming Ren, Yuxuan Sun, et~al. 2024.
\newblock {MMMU}: A massive multi-discipline multimodal understanding and reasoning benchmark for expert agi.
\newblock In \emph{Proceedings of the IEEE/CVF Conference on Computer Vision and Pattern Recognition}, pages 9556--9567.

\bibitem[{Zhang et~al.(2018)Zhang, Riesa, Gillick, Bakalov, Baldridge, and Weiss}]{zhang-etal-2018-fast-compact}
Yuan Zhang, Jason Riesa, Daniel Gillick, Anton Bakalov, Jason Baldridge, and David Weiss. 2018.
\newblock \href {https://doi.org/10.18653/v1/D18-1030} {A fast, compact, accurate model for language identification of codemixed text}.
\newblock In \emph{Proceedings of the 2018 Conference on Empirical Methods in Natural Language Processing}, pages 328--337, Brussels, Belgium. Association for Computational Linguistics.

\bibitem[{Zhu et~al.(2025)Zhu, Wang, Chen, Liu, Ye, Gu, Duan, Tian, Su, Shao et~al.}]{zhu2025internvl3}
Jinguo Zhu, Weiyun Wang, Zhe Chen, Zhaoyang Liu, Shenglong Ye, Lixin Gu, Yuchen Duan, Hao Tian, Weijie Su, Jie Shao, et~al. 2025.
\newblock {InternVL3}: Exploring advanced training and test-time recipes for open-source multimodal models.
\newblock \emph{arXiv preprint arXiv:2504.10479}.

\end{thebibliography}

\clearpage
\appendix

\section{Segment Categories in Dataset}
\subsection{Categories of Latin Usage}
\label{appendix:latin_categories}

We identified and annotated the following 12 categories of Latin usage found in the 18th-century documents:
\begin{enumerate}[topsep=4pt, partopsep=4pt]
 \itemsep-0.2em
    \item Bilingual Editions (Latin/English): Original Latin text and its English translation right next to it.
    \item Independent Latin Text: Original Latin text by the author, sometimes accompanied by English text on the same page.
    \item Direct Quotations in Latin: Latin phrases or sentences that are quoted verbatim, often within an otherwise predominantly English text.
    \item Code Switching: Text where the writer alternates between Latin and English within the same text, often for stylistic or rhetorical purposes.
    \item Dictionaries: Latin text appears in a dictionary like context, for example, with entries that define individual Latin words, often with translations or explanations in another language.
    \item Footnotes: Latin text appears in annotations or footnotes, often providing definitions or explanations for Latin words or phrases used in the main text.
    \item Emblematic quotes: Latin phrases or sentences are used as symbolic or thematic elements, often serving as mottos, epigraphs, or maxims. Typically, set apart from the main text, such as at the beginning of chapters, sections, or works.
    \item Sidenotes: Printed or authorial notes placed in the margins or alongside the main text.
    \item Legal Formulae: Standardized Latin phrases used in legal contexts.
    \item Ecclesiastical Formulae: Standardized Latin expressions used in religious or ecclesiastical contexts.
    \item Tables and Charts: Use of Latin in tabular data, genealogies, calendars, scientific diagrams, inflection tables, or mathematical charts.
    \item Indexes and Catalogs: Use of Latin in structured lists such as indices, bibliographies, or library catalogs.
\end{enumerate}

\subsection{Category Illustrations}
\label{appendix:append_data}

\begin{figure}[htb!]
    \centering
    \includegraphics[width=\linewidth, clip]{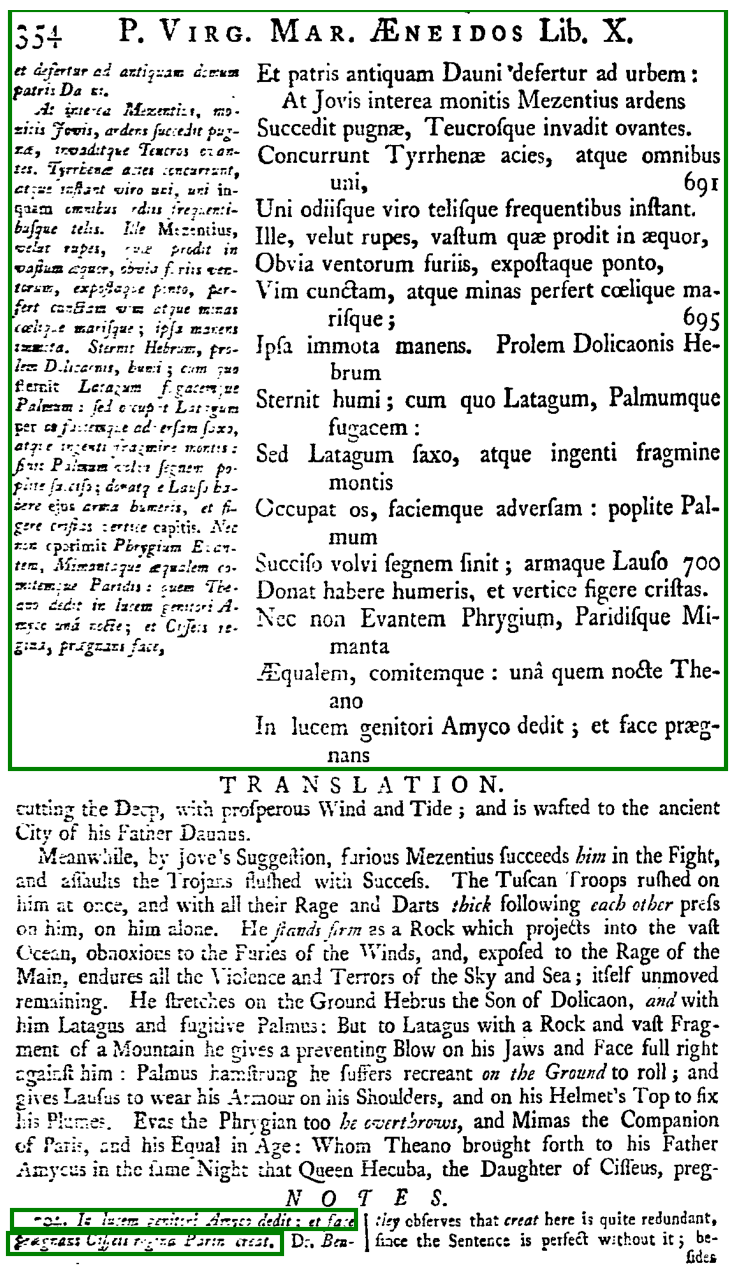}
    \caption{An example page with Latin fragments.}
    \label{fig:example-page2}
\end{figure}

\begin{figure}[htb!]
    \centering
    %  trim={<left> <lower> <right> <upper>}
    \includegraphics[width=\linewidth, clip]{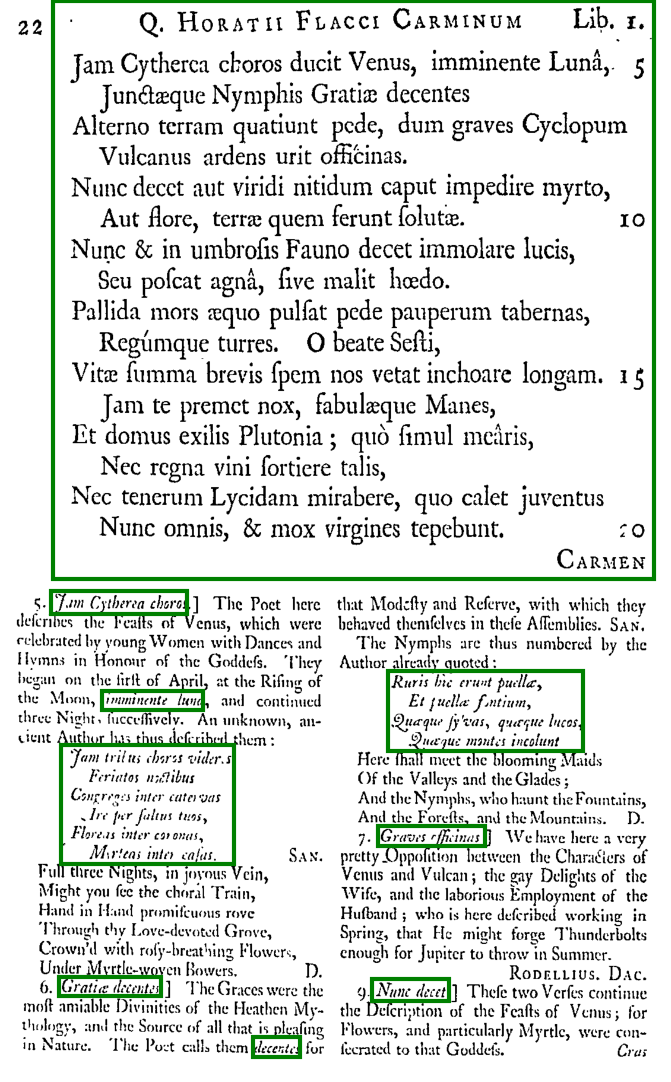}
    \caption{An example page with Latin fragments.}
    \label{fig:example-page3}
\end{figure}

Figures~\ref{fig:example-page2} and~\ref{fig:example-page3} show examples of Latin text categories. Both figures feature independent Latin text in the main text box at the top of the pages and footnotes at the bottom. Figure~\ref{fig:example-page2} is a bilingual edition of a Latin text, with an English translation directly below the Latin text at the top of the page. The footnotes in Figure~\ref{fig:example-page3} also include instances of code-switching and direct quotations of Latin text.


\section{Evaluation Setting Details}

\subsection{Preprocessing}
\label{appendix:preprocess}

This section details the text preprocessing pipeline for evaluation, implemented in Python, to normalize both ground truth and predicted text strings before unigram token extraction. The primary goal of this pipeline is to standardize textual representations, thereby mitigating the impact of superficial variations (e.g., from OCR noise or stylistic differences) on downstream metrics calculation. Note, that this applies to the evaluation step only, while Latin extracting models take an input text without these steps.  %All preprocessing steps are applied on a per-sentence basis to ensure that n-grams do not span across sentence boundaries.

For each text string, the following sequential operations are performed:

\begin{enumerate}
    % \item \textbf{Paragraph and Sentence Segmentation:}
    % \begin{itemize}
    %     \item Input texts are first split into paragraphs based on double newline characters. This step uses standard Python string operations.
    %     \item Each resulting paragraph is then further segmented into individual sentences using the \texttt{sent\_tokenize} function from the NLTK library (\texttt{nltk.tokenize}, version 3.9.1).
    % \end{itemize}
    % The subsequent preprocessing steps (2-8) are applied independently to each extracted sentence.

    \item \textbf{Unicode Normalization:}
    Each string undergoes Unicode normalization using the \texttt{normalize} with ``NFKD'' method from Python's built-in \texttt{unicodedata} module. This step decomposes characters into their canonical forms, for example, separating accents from base characters, which helps in standardizing character representation.

    \item \textbf{Ligature Replacement:}
    A predefined set of common ligatures is replaced with their constituent characters. Examples of replacements include `ﬀ' to \texttt{ff}, `æ' to \texttt{ae}, and importantly for some historical contexts, `\&' to \texttt{et}.

    \item \textbf{Lowercasing:}
    All alphabetic characters in the string are converted to lowercase.

    \item \textbf{Digit Removal:}
    All sequences of digits are removed from the string to avoid prediction ambiguity on digits, e.g., OCR digits in footnote notations. 

    \item \textbf{De-hyphenation (Word Merging):}
    This step addresses common OCR inconsistencies in handling end-of-line hyphens from historical documents. To ensure textual uniformity for subsequent analysis, word segments that were hyphenated, typically due to line breaks in the original source, are consistently merged into single tokens.

    \item \textbf{Punctuation Stripping:}
    All standard punctuation marks, as defined by Python's \texttt{string.punctuation} constant, are removed from the string.

    \item \textbf{Word Tokenization:}
    After the above cleaning steps, each processed sequence is tokenized into a list of individual words using the \texttt{word\_tokenize} function from the NLTK library (\texttt{nltk.tokenize}, version 3.9.1).

\end{enumerate}

\subsection{Fuzzy Matching Algorithm in Token-level Metrics}
\label{appendix:fuzzy_alg}

To evaluate segment correspondence, we apply a fuzzy matching algorithm to compare lists of preprocessed ground truth tokens against predicted tokens for each sample. This approach calculates Precision, Recall, and F1 score while being robust to minor textual variations. The core matching logic is outlined in Algorithm~\ref{alg:core_fuzzy_match}.

\begin{algorithm*}
\caption{Fuzzy Matching and Token Metrics Output}
\label{alg:core_fuzzy_match}
\begin{algorithmic}[1]
\Procedure{CalculateFuzzyMetrics}{$GT\_Tokens, Pred\_Tokens, \theta$}
    \State \Comment{Input: $GT\_Tokens$, $Pred\_Tokens$ (lists of preprocessed tokens)}
    \State \Comment{\phantom{Input:} $\theta$ (edit distance ratio threshold for a match)}
    \State \Comment{Output: Precision, Recall, F1 score}

    \State $TP \gets 0$
    \State $matched\_gt\_indices \gets \emptyset$

    \For{each $pred\_token$ in $Pred\_Tokens$}
        \For{each $gt\_token$ in $GT\_Tokens$ (with index $gt\_idx$)}
            \If{$gt\_idx \in matched\_gt\_indices$} \textbf{continue} \EndIf
            \If{\Call{IsFuzzyMatch}{$gt\_token, pred\_token, \theta$}}
                \State $TP \gets TP + 1$
                \State Add $gt\_idx$ to $matched\_gt\_indices$
                \State \textbf{break} \Comment{Current $pred\_token$ matched}
            \EndIf
        \EndFor
    \EndFor

    \State $FP \gets \text{length}(Pred\_Tokens) - TP$
    \State $FN \gets \text{length}(GT\_Tokens) - TP$

    \State $Precision \gets TP / (TP + FP)$
    \State $Recall \gets TP / (TP + FN)$ 
    \State $F1 \gets 2 \times (Precision \times Recall) / (Precision + Recall)$ 
    \State \Return $Precision, Recall, F1$
\EndProcedure
\end{algorithmic}
\end{algorithm*}

The algorithm performs a greedy, one-to-one fuzzy match: each predicted token is compared against available ground truth tokens using a match indicator function (\texttt{IsFuzzyMatch}) based on edit distance and a predefined proportion threshold $\theta$. A match only holds when the edit distance is less than or equal to $\theta$ proportion of the length of the ground truth token string. A ground truth token can only be matched once to ensure an accurate count of distinct true positive matches. This fuzzy approach is beneficial as it offers robustness to minor textual variations that may persist even after preprocessing, leading to a more meaningful evaluation of segment correspondence.

\section{Main Prompt Details}
\label{appendix:prompts}

This section details the exact prompt templates employed to instruct the LLMs for the main experiment of Latin script detection and extraction. The prompts were adapted based on the input modality being used. In the templates below, the placeholder \texttt{\{page\_text\}} indicates where the OCR output corresponding to the processed page image was dynamically inserted.

\paragraph{Image + Text}
%The following prompt was used when providing the model with both the page image and its corresponding OCR text:
\begin{quote}
\ttfamily % Use typewriter font for the entire prompt
Identify and extract all segments written in Latin (e.g., Classical or Medieval Latin) from the provided image, using the accompanying OCR text as a reference. Return the results as a list of strings in the JSON format: \texttt{[``text1'', ``text2'', ...]}.

\medskip % Adds some vertical space, equivalent to a paragraph break
\noindent OCR Text: \texttt{\{page\_text\}}
\end{quote}

\paragraph{Image-only}
%When only the page image was provided to the model, the following prompt was used:
\begin{quote}
\ttfamily
Identify and extract all segments written in Latin (e.g., Classical or Medieval Latin) from the provided image. Return the results as a list of strings in the JSON format: \texttt{[``text1'', ``text2'', ...]}.
\end{quote}

\paragraph{Text-only}
%For models that only accepted textual input, the following prompt, incorporating the OCR text, was used:
\begin{quote}
\ttfamily
Identify and extract all segments written in Latin (e.g., Classical or Medieval Latin) from the OCR text of an image. Return the results as a list of strings in the JSON format: \texttt{[``text1'', ``text2'', ...]}.

\medskip % Adds some vertical space
\noindent OCR Text: \texttt{\{page\_text\}}
\end{quote}

\section{Configuration of Baseline}
\label{appendix:baseline}

For comparative language identification, we employed Lingua (version 2.1.0) \citep{pemistahl2021lingua} as a baseline. The \texttt{LanguageDetector} was specifically configured to operate with a predefined restricted set of eight languages: English, French, German, Greek, Italian, Spanish, Portuguese, and Latin. This selection aims to encompass Latin itself and a set of the most frequently occurring languages within our target corpus ECCO (English, French, German, and Greek), while also including languages present in ECCO that share orthographic or lexical similarities with Latin (Italian, Spanish, and Portuguese). Including these similar languages was intended to create a more global and robust test scenario for accurate Latin identification in ECCO. 

In our pipeline, Lingua's function to detect multiple languages within a given text (\texttt{detect\_multiple\_languages\_of} method) was utilized on each page's OCR output. From the resulting language segments identified by Lingua, only those substrings classified as Latin were subsequently extracted for our analysis and evaluation.

\section{Additional Results}
\label{appendix:additional_results}

\subsection{Categorization Task Results}
\label{appendix:cat_res}

This section details the design and results of our complementary joint extraction and categorization experiment stated in Section~\ref{sec:cat_res}, a more demanding task created to assess the models' deeper, functional understanding of text. The following subsections are structured to first outline the task's setup, including the specific prompt and the evaluation protocol. Following this methodological overview, we will present and analyze the detailed performance of the top models to highlight their capabilities and limitations on this challenge.

\paragraph{Task Design}
To move beyond simple Latin text retrieval and directly assess the models' ability to comprehend the functional role of Latin segments, we designed a more challenging joint extraction and categorization task. In this task, a model is required not only to identify and extract all Latin text from a given page but also to simultaneously assign each extracted segment to one of twelve predefined functional categories. The required output is a structured JSON object where keys correspond to the predefined category names and the values are lists of text segments assigned to each category. This task design compels the model to make explicit judgments about the semantic and contextual purpose of the text, thereby providing a clearer signal of its deeper comprehension abilities than a simple extraction task would require.

\paragraph{Task Prompt and Evaluation Methodology}
The core prompt for the joint task is slightly adapted based on the model's input modality (e.g., text-only, image-only, or multimodal). The version shown below is for the multimodal (I+T) setting:

\begin{quote}
\ttfamily % Use typewriter font for the entire prompt
Identify and extract all segments written in Latin (e.g., Classical or Medieval Latin) from the provided image.

\medskip
\noindent After extracting all the Latin segments, assign each to one or some of the following categories:
\begin{itemize}[leftmargin=*,label=\texttt{-}]
    \item \textbf{Bilingual}: Original Latin text with its English translation immediately following.
    \item \textbf{Independent}: Standalone Latin text by the author, possibly with adjacent English.
    \item \textbf{Direct Quote}: Latin text quoted verbatim within primarily English content.
    \item \textbf{Code-switching}: Alternation between Latin and English within the same passage.
    \item \textbf{Dictionary}: Latin entries in dictionary-style definitions or explanations.
    \item \textbf{Footnote}: Latin in annotations or footnotes clarifying main text.
    \item \textbf{Emblematic}: Latin used as mottos, epigraphs, or thematic standalone phrases.
    \item \textbf{Side-note}: Marginal notes or annotations in Latin beside main text.
    \item \textbf{Legal}: Standard Latin phrases used in legal contexts.
    \item \textbf{Ecclesiastical}: Standard Latin phrases used in religious contexts.
    \item \textbf{Tables and Charts}: Latin in tables, charts, genealogies, calendars, scientific or inflection data.
    \item \textbf{Indices and Catalogs}: Latin in lists, indices, bibliographies, or catalog entries.
\end{itemize}

\medskip
\noindent Return a JSON object mapping each category to a list of Latin text segments, using exactly this format (no extra text or modifications):
\texttt{\{``Bilingual'': [...], ``Independent'': [...], ``Direct Quote'': [...], ``Code-switching'': [...], ``Dictionary'': [...], ``Footnote'': [...], ``Emblematic'': [...], ``Side-note'': [...], ``Legal'': [...], ``Ecclesiastical'': [...], ``Tables and Charts'': [...], ``Indices and Catalogs'': [...]\}}

\medskip
\noindent If a category has no results, include it with an empty list.

\medskip
\noindent OCR Text: \texttt{\{page\_text\}}
\end{quote}

Evaluation for this joint task is performed on a per-category basis. For each of the twelve categories, the list of text segments returned by the model under that category's key in the JSON output is treated as the predicted set. This set is then compared against the ground-truth list of segments annotated for that same category.

Precision, Recall, and F1 scores are then calculated at the token level for each category independently, using the same fuzzy matching process described in Section~\ref{sec:metrics}. The final Macro F1 score, reported in Table~\ref{tab:cat_anno_results}, is the unweighted arithmetic mean of these twelve individual F1 scores. This method allows us to assess not only the model's overall performance but also its specific strengths and weaknesses with respect to the understanding of each functional type of Latin text.

\paragraph{Results and Analysis}

Table~\ref{tab:cat_anno_results} presents the per-category F1 (F), Precision (P), and Recall (R) scores for three top-performing models. As noted in our main text, the overall performance on this demanding task is terrible, with the best model, DeepSeek-R1, achieving a Macro F1 score of only 20.98\%.

Several key observations can be drawn from these results. Firstly, the text-only DeepSeek-R1 significantly outperforms the multimodal Qwen-VL models, suggesting that the underlying language model's reasoning capability, rather than visual cues, is the current dominant factor for this specific categorization task. Secondly, there is a drastic variance in performance across categories. Notably, all models completely fail on the ``Side-note'' category, each scoring an F1 of 0.00, although it should have strong visual layout evidence. Performance is also exceptionally poor on categories requiring high contextual awareness, such as ``Code-switching'' and ``Dictionary'' entries. The relatively highest scores are achieved on categories with more distinct and self-contained structures, like ``Direct Quote'' and ``Independent'' sections, though even here the best F1 scores remain below 45\%.

The overall low scores and high variance across categories underscore the challenge faced by current LLMs. Specifically, performance appears to be heavily influenced by the models' intrinsic biases, likely reflecting the training data distribution. The models exhibit partial capability on common, text-centric categories like ``Direct Quote'' but almost completely fail to classify rarer or more specialized, layout-dependent categories such as ``Side-note''. This area warrants significant further investigation as a dedicated future work. Focused research should first aim to diagnose the primary failure modes more precisely. For instance, future studies could seek to disentangle the core classification challenge from the requirement of generating structured joint output, and to determine whether poor performance stems from a fundamental lack of historical knowledge, poor text grounding in noisy document images, or suboptimal prompt design. Answering these foundational questions is a necessary next step before exploring more complex interventions like specialized training or novel model architectures.

\begin{sidewaystable}[p]
\setlength{\tabcolsep}{2pt}
\centering
\begin{adjustbox}{max width=\textheight}
\begin{tabular}{c|ccc|ccc|ccc|ccc|ccc|ccc|ccc|ccc|ccc|ccc|ccc|ccc|c}
\toprule
\multirow{2}{*}{\textbf{Model}}
& \multicolumn{3}{c|}{\textit{Bili}} 
& \multicolumn{3}{c|}{\textit{Code}} 
& \multicolumn{3}{c|}{\textit{Dict}} 
& \multicolumn{3}{c|}{\textit{Quote}} 
& \multicolumn{3}{c|}{\textit{Eccl}} 
& \multicolumn{3}{c|}{\textit{Embl}} 
& \multicolumn{3}{c|}{\textit{Foot}} 
& \multicolumn{3}{c|}{\textit{Indep}} 
& \multicolumn{3}{c|}{\textit{Index}} 
& \multicolumn{3}{c|}{\textit{Legal}} 
& \multicolumn{3}{c|}{\textit{Side}} 
& \multicolumn{3}{c|}{\textit{Table}} 
& \multirow{2}{*}{\textbf{Macro F1}} \\
\cline{2-37}
& \textbf{F} & \textbf{P} & \textbf{R} 
& \textbf{F} & \textbf{P} & \textbf{R} 
& \textbf{F} & \textbf{P} & \textbf{R} 
& \textbf{F} & \textbf{P} & \textbf{R} 
& \textbf{F} & \textbf{P} & \textbf{R} 
& \textbf{F} & \textbf{P} & \textbf{R} 
& \textbf{F} & \textbf{P} & \textbf{R} 
& \textbf{F} & \textbf{P} & \textbf{R} 
& \textbf{F} & \textbf{P} & \textbf{R} 
& \textbf{F} & \textbf{P} & \textbf{R} 
& \textbf{F} & \textbf{P} & \textbf{R} 
& \textbf{F} & \textbf{P} & \textbf{R} &
\\
\midrule
DeepSeek-R1 (T) 
& 18.82 & 19.81 & 18.48 
& 3.16 & 4.50 & 3.07 
& 6.51 & 7.58 & 6.45 
& 43.01 & 46.44 & 43.00 
& 27.39 & 28.29 & 28.11 
& 16.28 & 16.47 & 16.16 
& 30.11 & 36.03 & 28.99 
& 44.71 & 47.13 & 44.21 
& 25.38 & 29.26 & 24.02 
& 23.84 & 26.99 & 23.25 
& 0.00 & 0.00 & 0.00 
& 12.49 & 17.84 & 10.86 
& 20.98 \\
Qwen2.5-VL-32B (I+T) 
& 18.13 & 19.07 & 17.86 
& 0.23 & 0.13 & 0.77 
& 6.04 & 7.25 & 5.59 
& 32.10 & 35.30 & 31.78 
& 14.28 & 19.80 & 13.01 
& 16.36 & 16.55 & 16.19 
& 18.70 & 22.27 & 20.41 
& 29.17 & 32.97 & 27.99 
& 12.26 & 20.04 & 9.68 
& 13.02 & 22.56 & 11.68 
& 0.00 & 0.00 & 0.00 
& 15.19 & 21.71 & 13.71 
& 14.62 \\
Qwen2.5-VL-32B (I) 
& 16.15 & 17.77 & 15.78 
& 0.00 & 0.00 & 0.00 
& 3.56 & 4.72 & 2.86 
& 25.81 & 30.13 & 25.23 
& 9.61 & 21.13 & 8.48 
& 7.00 & 7.04 & 6.97 
& 13.24 & 16.47 & 13.68 
& 28.60 & 31.14 & 28.26 
& 2.28 & 5.83 & 1.45 
& 8.38 & 18.76 & 7.16 
& 0.00 & 0.00 & 0.00 
& 3.88 & 6.90 & 2.70 
& 9.88 \\
\bottomrule
\end{tabular}
\end{adjustbox}
\caption{Categorization performance of the best models, where Macro F1 denotes the unweighted average of category-wise F1 scores. Category names in the header are abbreviated for brevity (e.g., \textit{Bili.} for \textit{Bilingual}).}
\label{tab:cat_anno_results}
\end{sidewaystable}

\subsection{Prompt Experiments}
\label{appendix:prompt_exp}

This section provides the full content of the prompt strategies used in our prompt engineering experiments, along with a table of their complete result numbers.

The following are the specific instructions given to the Qwen2.5-VL-32B model for each prompt strategy. The base prompt shown in Appendix~\ref{appendix:prompts}, used for the \textbf{Minimal} strategy, forms the core of most other prompts. For brevity, the final line of the prompt providing the OCR text input, \texttt{OCR Text: \{page\_text\}}, is omitted from each example below as its format is consistent across all experiments.

\paragraph{Partial Categories}
\begin{quote}
\ttfamily
Identify and extract all segments written in Latin (e.g., Classical or Medieval Latin) from the provided image, using the accompanying OCR text as a reference.  

Return the results as a list of strings in the JSON format: [``text1'', ``text2'', ...]. 

\color{purple}{Pay particular attention to identifying Latin segments in code-switching, dictionaries, footnotes, sidenotes, tables, and charts, while maintaining accuracy across all other categories.}
\end{quote}

\paragraph{All Categories}
\begin{quote}
\ttfamily
Identify and extract all segments written in Latin (e.g., Classical or Medieval Latin) from the provided image, using the accompanying OCR text as a reference.  

Return the results as a list of strings in the JSON format: [``text1'', ``text2'', ...]. 

\color{purple}{Please pay attention to Latin in all of those categories: Bilingual, Independent, Direct Quote, Code-switching, Dictionary, Footnote, Emblematic, Side-note, Legal, Ecclesiastical, Tables and Charts, Indices and Catalogs.}
\end{quote}

\paragraph{Detailed Categories}
\begin{quote}
\ttfamily
Identify and extract all segments written in Latin (e.g., Classical or Medieval Latin) from the provided image, using the accompanying OCR text as a reference. 

Return the results as a list of strings in the JSON format: [``text1'', ``text2'', ...]. 

{\color{purple}Please pay attention to Latin in all of those categories: 

- Bilingual: Original Latin text with its English translation immediately following. 

- Independent: Standalone Latin text by the author, possibly with adjacent English. 

- Direct Quote: Latin text quoted verbatim within primarily English content. 

- Code-switching: Alternation between Latin and English within the same passage. 

- Dictionary: Latin entries in dictionary-style definitions or explanations. 

- Footnote: Latin in annotations or footnotes clarifying main text. 

- Emblematic: Latin used as mottos, epigraphs, or thematic standalone phrases. 

- Side-note: Marginal notes or annotations in Latin beside main text. 

- Legal: Standard Latin phrases used in legal contexts. 

- Ecclesiastical: Standard Latin phrases used in religious contexts. 

- Tables and Charts: Latin in tables, charts, genealogies, calendars, scientific or inflection data. 

- Indices and Catalogs: Latin in lists, indices, bibliographies, or catalog entries.}
\end{quote}

\paragraph{Specialist Context}
\begin{quote}
\ttfamily
{\color{purple}You are a specialist in classical languages and historical documents. You are given a scanned image of a page from an 18th-century document and its corresponding OCR result.}

\ttfamily
Identify and extract all segments written in Latin (e.g., Classical or Medieval Latin) from the provided image, using the accompanying OCR text as a reference.  

Return the results as a list of strings in the JSON format: [``text1'', ``text2'', ...].
\end{quote}

\paragraph{Specialist}
\begin{quote}
{\ttfamily

{\color{purple}You are a specialist in classical languages and historical documents.}

Identify and extract all segments written in Latin (e.g., Classical or Medieval Latin) from the provided image, using the accompanying OCR text as a reference.  

Return the results as a list of strings in the JSON format: [``text1'', ``text2'', ...].}
\end{quote}

\paragraph{Metrics}
\begin{quote}
\ttfamily
Identify and extract all segments written in Latin (e.g., Classical or Medieval Latin) from the provided image, using the accompanying OCR text as a reference.  

Return the results as a list of strings in the JSON format: [``text1'', ``text2'', ...]. 

\color{purple}{Please ensure your extraction is both precise (no non-Latin segments are included) and comprehensive (all Latin segments are found).}
\end{quote}

\paragraph{Empty List}
\begin{quote}
\ttfamily
Identify and extract all segments written in Latin (e.g., Classical or Medieval Latin) from the provided image, using the accompanying OCR text as a reference.  

Return the results as a list of strings in the JSON format: [``text1'', ``text2'', ...]. 

\color{purple}{Return an empty list if no Latin text is found.}
\end{quote}

\paragraph{Single Word}
\begin{quote}
\ttfamily
Identify and extract all segments written in Latin (e.g., Classical or Medieval Latin) from the provided image, using the accompanying OCR text as a reference.  

{\color{purple}Include segments even if they consist of only a single Latin word. }

Return the results as a list of strings in the JSON format: [``text1'', ``text2'', ...].
\end{quote}

\paragraph{No Abbrev}
\begin{quote}
\ttfamily
Identify and extract all segments written in Latin (e.g., Classical or Medieval Latin) from the provided image, using the accompanying OCR text as a reference.  

Return the results as a list of strings in the JSON format: [``text1'', ``text2'', ...]. 

\color{purple}{Please do not include any abbreviations that are commonly used in contemporary languages, such as "etc.", "e.g.", "i.e.", "et al.", "a.m.", "p.m.", "A.D.", "B.C.", "P.S.", and similar.}
\end{quote}

\paragraph{No Borrow}
\begin{quote}
\ttfamily
Identify and extract all segments written in Latin (e.g., Classical or Medieval Latin) from the provided image, using the accompanying OCR text as a reference. 

Return the results as a list of strings in the JSON format: [``text1'', ``text2'', ...]. 

\color{purple}{Please consider the language context and do not include Latin words or phrases that are used as loanwords or integrated into other languages, unless they function as Latin text in the context.}
\end{quote}

\begin{table}[htb!]
\centering
\begin{adjustbox}{max width=0.45\textwidth}
\begin{tabular}{c|ccc|ccc}
\toprule
\multirow{2}{*}{\textbf{Prompt Strategy}} & \multicolumn{3}{c|}{\textsc{Page-Level}} & \multicolumn{3}{c}{\textsc{Token-Level}} \\
\cline{2-7}
 & \textbf{F} & \textbf{P} & \textbf{R} & \textbf{F} & \textbf{P} & \textbf{R} \\
\midrule
Minimal & 96.18 & 92.94 & 99.66 & 84.32 & 86.90 & 83.99 \\
Partial Categories & 96.27 & 92.95 & 99.83 & 84.90 & 85.35 & \textbf{86.61} \\
All Categories & 96.55 & 94.23 & 98.99 & 84.20 & 85.53 & 85.01 \\
Detailed Categories & 96.41 & 93.65 & 99.33 & 83.63 & 84.72 & 84.63 \\
Specialist Context & 95.35 & 91.10 & \textbf{100.00} & 84.79 & 87.12 & 84.85 \\
Specialist & \textbf{96.81} & \textbf{94.26} & 99.49 & 84.47 & 86.74 & 84.28 \\
Metrics & 94.56 & 90.09 & 99.49 & 84.40 & 87.19 & 83.75 \\
Empty List & 96.06 & 93.75 & 98.48 & 83.84 & 86.60 & 83.03 \\
Single Word & 92.36 & 86.05 & 99.66 & 84.54 & 86.02 & 84.90 \\
No Abbrev & 95.01 & 91.05 & 99.33 & 84.85 & 87.23 & 84.50 \\
No Borrow & 96.41 & 93.51 & 99.49 & \textbf{85.09} & \textbf{88.04} & 84.02 \\
\bottomrule
\end{tabular}
\end{adjustbox}
\caption{Impact of prompting on Qwen2.5-VL-32B.}
\label{tab:ablation}
\end{table}

As summarized in the main text, most prompt variations result in a precision-recall trade-off. Table~\ref{tab:ablation} provides the detailed metric numbers. Notably, adopting a ``Specialist'' persona yields the highest overall page-level performance (F1 96.81). Interestingly, adding more specific situational details to this persona (``Specialist Context'') achieves perfect page-level recall (100.00\%) but at the cost of significantly lower precision, resulting in a lower F1 score. This highlights the delicate balance in prompt design. Furthermore, simply providing more instructions on specialized knowledge, such as in the ``All Categories'' and ``Detailed Categories'' prompts, does not guarantee a significant improvement over the ``Minimal'' baseline, indicating that a detailed and informative prompt is not directly correlated with better performance. This reveals that a core deficit in contextual understanding remains the primary bottleneck.

For the token-level task, negative constraints like in the ``No Borrow'' and ``No Abbrev'' prompts yield the higher F1 scores (85.09 and 84.85 respectively), primarily by increasing precision. This indicates that providing a more precise, linguistically-grounded definition of the task is moderately effective. Ideally, however, a model should possess this specialist language identification capability intrinsically, distinguishing true Latin from common borrowings or abbreviations without requiring such explicit constraints. This requires moving beyond simple etymological recognition to a pragmatic understanding of a word's \textit{function}, enabling the model to differentiate between genuine code-switching into Latin and fully assimilated loanwords (e.g., ``status quo'') or abbreviations (e.g., ``et al.''). The fact that explicit negative constraints are needed even to approximate this behavior highlights a key limitation of current models. It reveals that they still rely on manually encoded rules to navigate nuanced linguistic boundaries that a human expert would discern implicitly. Achieving this level of intrinsic, context-sensitive discernment remains a significant and challenging long-term goal for the development of truly knowledgeable AI agents as domain experts.

% todo: add more references
\subsection{Fuzzy Matching Threshold}
\label{appendix:theta}

\begin{figure*}[htb!]
    \centering
    \includegraphics[width=\textwidth]{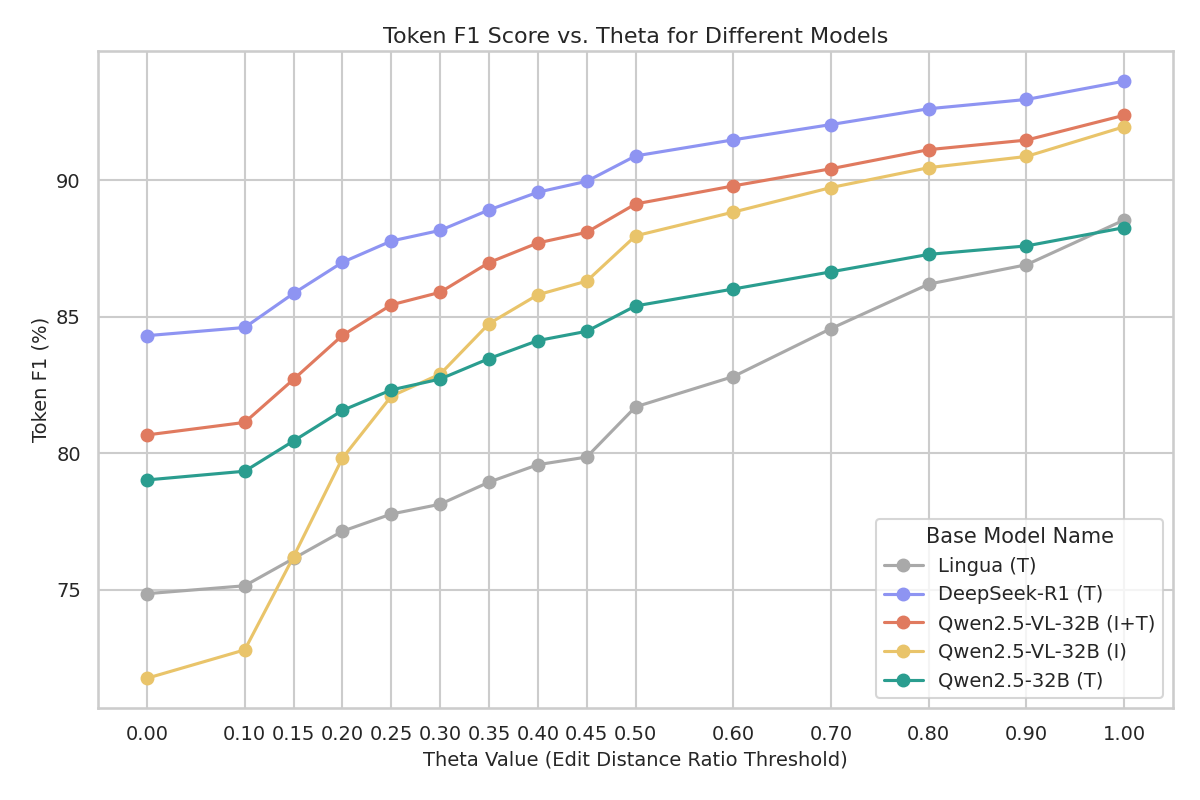}
    \caption{Token F1 scores on different $\theta$ value.}
    \label{fig:theta}
\end{figure*}

The fuzzy matching threshold, $\theta$ (representing the maximum allowed normalized edit distance relative to ground truth token length), was empirically set to 0.2 for all the experiments. This choice aligns with a common heuristic of tolerating approximately ``1 error in 5 characters,'' suitable for OCR-derived text, and is supported by our sensitivity analysis in Figures~\ref{fig:theta}. It consistently shows that while F1 scores generally increase with $\theta$, the most substantial and steepest F1 score improvements for the majority of evaluated models are concentrated in the range leading up to $\theta \approx 0.2$, effectively compensating for common, fine-grained textual variations attributable to OCR noise. Although metrics may continue to rise beyond this point for some configurations on our dataset, we maintain $\theta=0.2$ as a principled trade-off. A higher universal threshold could risk over-tolerating more substantial prediction errors beyond typical OCR noise, potentially prioritizing the matching of token quantity or approximate form over precise content fidelity. This could also obscure true output quality differences, especially when comparing models with varying input noise levels (e.g., image-only versus OCR-input systems).

\subsection{Qualitative Results}
\label{appendix:qualitative}
More qualitative results are shown as examples to illustrate the best model's performance and the error modes.

\begin{figure*}[p]
    \centering
    \includegraphics[width=\textwidth]{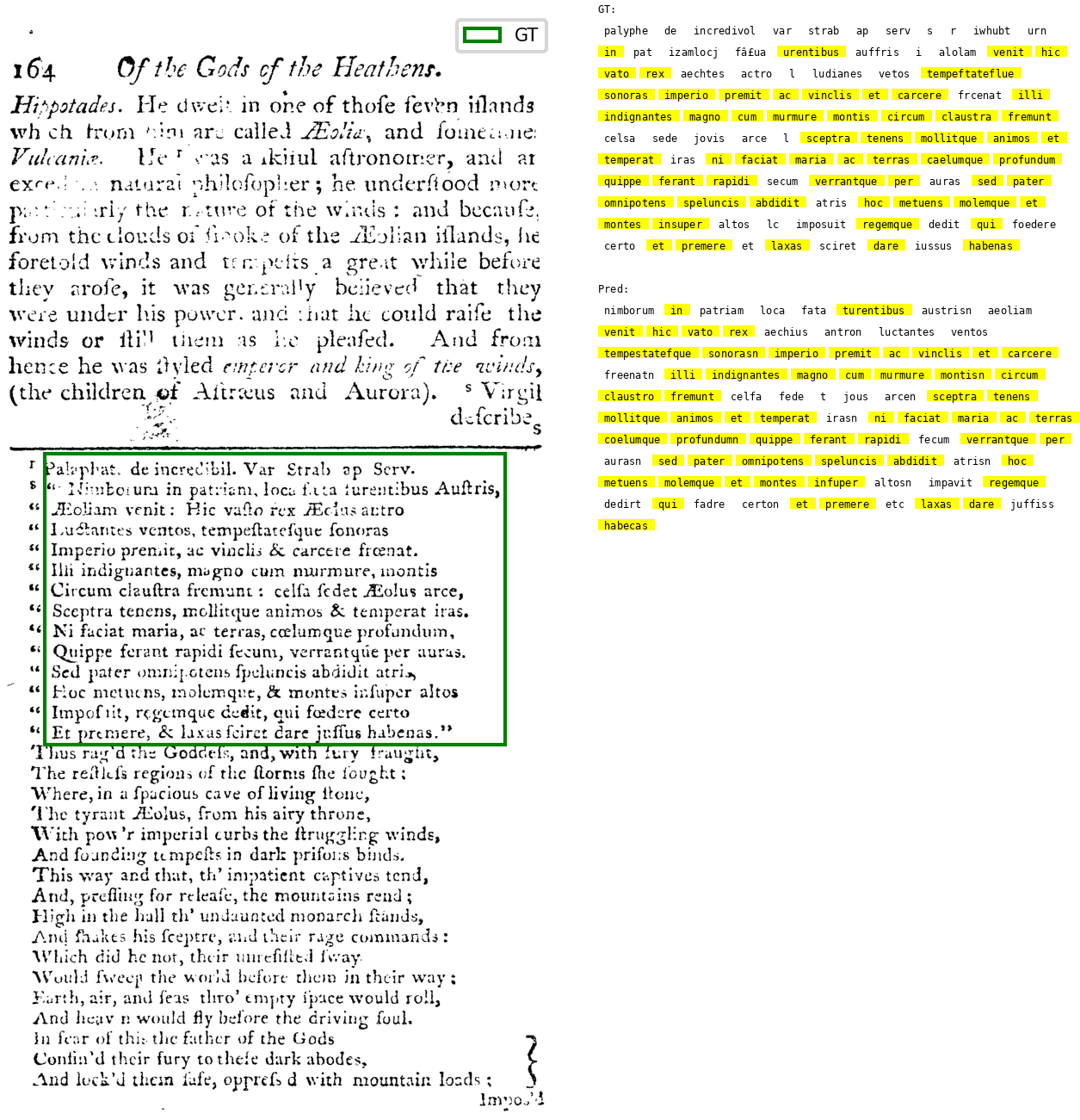}
    \caption{An example page with Latin fragments, together with our ground truth and prediction for that page.}
    \label{fig:example-page4}
\end{figure*}

Figure~\ref{fig:example-page4} shows an example of a mismatch caused by significant OCR noise caused by poor original image quality. Here, the post-corrected OCR of our ground truth differs so much from the OCR visual or MLLMs produced during the prediction process that not even our edit distance-based fuzzy ground truth matching can recover what is essentially a full match. This kind of error especially affects pages in the footnotes, code switching and dictionary categories, since the Latin texts in these categories tend to be printed in harder to detect fonts and layouts, which are additionally more likely to be affected by bad scan quality.

\begin{figure*}[p]
    \centering
    \includegraphics[width=\textwidth]{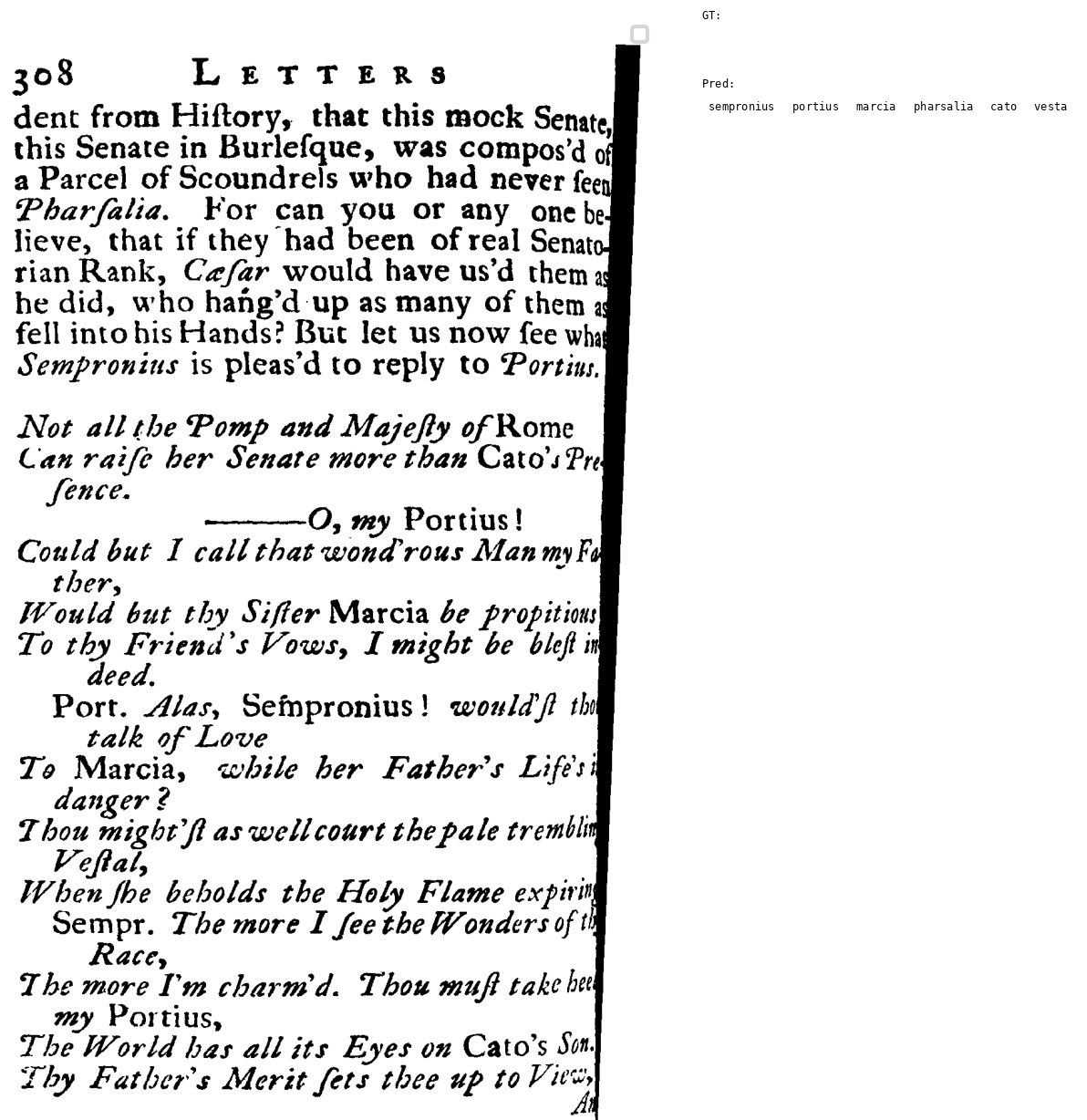}
    \caption{An example page without Latin fragments, together with our ground truth and prediction for that page.}
    \label{fig:example-page5}
\end{figure*}

Figure~\ref{fig:example-page5} shows an example of a definitional misunderstanding of the predictions, which is the typical phenomenon also discussed in the prompt experiment in Section~\ref{sec:prompts}. Although there is no Latin text on the page, the prediction contains the Roman names appearing in the page text.

\begin{figure*}[p]
    \centering
    \includegraphics[width=\textwidth]{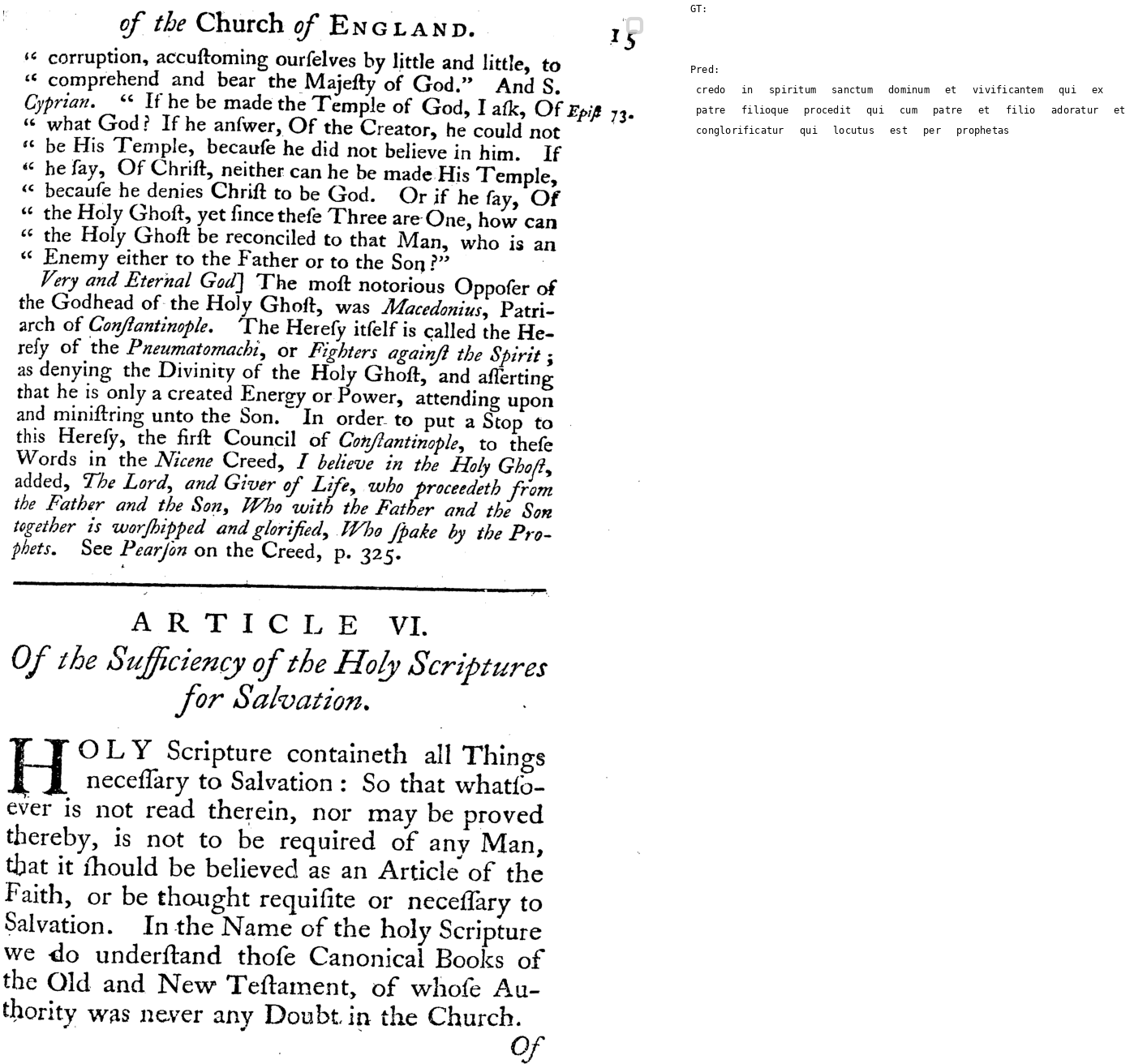}
    \caption{An example page without Latin fragments, together with our ground truth and hallucinated prediction for that page.}
    \label{fig:example-page6}
\end{figure*}

Figure~\ref{fig:example-page6} shows an example of a page where the prediction contains hallucinations. The model took part of the text and translated it into Latin in the prediction, without being prompted to do so.

\clearpage

\end{document}